\documentclass[preprint,11pt,3p,number]{elsarticle}

\usepackage{amsmath,amssymb}

\usepackage{algorithm}
\usepackage{algpseudocode}

\usepackage{graphicx}
\usepackage{booktabs}
\usepackage{longtable}
\usepackage{tabularx}
\usepackage{subcaption}
\usepackage{array}

\usepackage{float}
\usepackage{placeins}
\usepackage{setspace}
\usepackage{chngcntr}

\usepackage{appendix}

\usepackage{xcolor}

\biboptions{sort&compress}
\usepackage{hyperref}
\usepackage{cleveref}

\begin{document}
\doublespacing

\begin{frontmatter}

\title{A Hybrid GNN--FEM Framework for Phase-Field Fracture Simulation: Physics-Preserving Hybridization for Generalizable Surrogate Modeling}

\author[label1]{Hyeonbin Moon\fnref{fn1}}
\author[label2]{Yongjin Choi\fnref{fn1}}
\author[label1,label2,label4]{Seunghwa Ryu\corref{cor1}}

\fntext[fn1]{These authors contributed equally to this work.}
\cortext[cor1]{Corresponding author}
\ead{ryush@kaist.ac.kr}

\affiliation[label1]{
organization={Department of Mechanical Engineering, Korea Advanced Institute of Science and Technology (KAIST)},
city={Daejeon},
postcode={34141}, 
country={Republic of Korea}
}

\affiliation[label2]{
organization={KAIST InnoCORE PRISM-AI Center, Korea Advanced Institute of Science and Technology (KAIST)},
city={Daejeon},
postcode={34141}, 
country={Republic of Korea}
}

\affiliation[label4]{
organization={Department of AX, College of AI, Korea Advanced Institute of Science and Technology (KAIST)},
city={Daejeon},
postcode={34141},
country={Republic of Korea}
}

\begin{singlespace}
\begin{abstract}

Scientific machine learning (SciML) has emerged as a promising approach for accelerating simulations of complex physical systems, yet achieving physically consistent and generalizable predictions for nonlinear, history-dependent problems remains a central challenge. In this study, we propose a hybrid GNN--FEM framework for efficient and generalizable phase-field fracture modeling. While phase-field approaches provide a robust variational framework for simulating complex crack evolution, their high computational cost limits practical applications due to the need to solve a coupled, nonlinear, and history-dependent system within an incremental finite element (FEM) procedure. To address this challenge, a graph neural network (GNN) surrogate is integrated into the conventional staggered scheme, replacing the phase-field update at each load increment while retaining the FEM-based displacement solver to enforce mechanical equilibrium and boundary conditions. By preserving the incremental solution structure, the framework maintains consistency with the underlying history-dependent fracture evolution without requiring the surrogate to approximate the full solution trajectory. This selective surrogate strategy emphasizes the identification of a physically meaningful and incrementally structured learning target, rather than relying on brute-force data generation to directly learn the full fracture trajectory. The proposed framework achieves strong generalization across varying geometries, loading conditions, material properties, and discretizations beyond the training configuration, through dimensionless feature design, a graph-based formulation on mesh-based domains, and a physics-informed loss based on the governing phase-field equation. Numerical experiments demonstrate that the hybrid approach reduces computational cost while maintaining accuracy compared to conventional FEM, and exhibits robust predictive performance across a wide range of problem settings.

\end{abstract}
\end{singlespace}

\begin{singlespace}
\end{singlespace}

\end{frontmatter}

\clearpage

\pdfstringdefDisableCommands{\let\mathbf\relax\let\mathcal\relax\let\boldsymbol\relax}
{\noindent\bfseries Nomenclature}
\vspace{4pt}
{\setlength{\extrarowheight}{0pt}%
\renewcommand{\arraystretch}{0.82}%
\begin{longtable}{>{\raggedright\arraybackslash}p{0.25\textwidth} >{\raggedright\arraybackslash}p{0.68\textwidth}}
\toprule
Symbol & Description  \\
\midrule
\endhead
\midrule
\multicolumn{2}{r}{Continued on next page}\\
\endfoot
\bottomrule
\endlastfoot

$\Omega$, $\partial\Omega$ & Body domain and its boundary \\
$\mathbf{u}$ & Displacement field \\
$d$ & Crack phase-field ($0$ = intact, $1$ = fully cracked) \\
$\ell_c$ & Regularization length scale of the diffusive crack \\
$G_c$ & Critical energy release rate  \\
$g(d)$ & Degradation function for damaged stiffness \\
$\boldsymbol{\varepsilon}(\mathbf{u})$ & Infinitesimal strain tensor \\
$\psi_0(\boldsymbol{\varepsilon}),\;\psi_0^{+},\;\psi_0^{-}$ & Elastic strain-energy density (total, tensile, compressive part) \\
$\boldsymbol{\sigma}(\mathbf{u},d)$ & Damaged Cauchy stress tensor \\
$H(\mathbf{x},t)$ & History-dependent crack-driving field \\
$\lambda,\;\mu$ & Lam\'{e} constants \\
$E,\;\nu$ & Young's modulus and Poisson's ratio \\[3pt]

$\mathcal{T}_h,\;h$ & Finite element mesh and characteristic mesh size \\
$\mathbf{u}_n,\;d_n,\;H_n$ & Displacement, phase-field, and history field at step $n$ \\
$\mathbf{R}_u,\;\mathbf{R}_d$ & Displacement and phase-field residual vectors \\[3pt]

$\mathcal{G}=(\mathcal{V},\mathcal{E})$ & Mesh-induced graph (nodes $\mathcal{V}$, edges $\mathcal{E}$) \\
$\mathbf{x}_i$ & Node-feature vector at node $i$ \\
$H/(G_c\ell_c)$ & Normalized history field \\
$\mathbf{e}_{ij}$ & Edge-feature vector for edge $(i,j)$ \\
$\hat{d}_i$ & Predicted phase-field value at node $i$ \\
$H_{\mathrm{lat}},\;K$ & Latent feature dimension and number of message-passing iterations \\
$\mathbf{z}_i^{(k)},\;\mathbf{f}_{ij}^{(k)}$ & Latent node and edge representations at iteration $k$ \\
$\mathcal{N}(i)$ & Neighbour set of node $i$ \\
$\mathrm{MLP}_{\mathrm{node/edge/dec}}$ & Encoder and decoder multilayer perceptrons \\[3pt]

$\mathcal{L}_{\mathrm{data}},\;\mathcal{L}_{\mathrm{phys}}$ & Data-driven loss and physics-informed loss \\
$\lambda_{\mathrm{phys}}$ & Weighting coefficient for the physics-informed loss \\[3pt]

\end{longtable}}%

\section{Introduction}

Scientific machine learning (SciML) has emerged as a prominent research direction for modeling complex physical systems, in which data-driven surrogate models are constructed to approximate solutions of partial differential equations (PDEs) 
\cite{brunton2024promising,azizzadenesheli2024neural,li2020fourier,lu2021learning}. 
Among the various approaches within this field, physics-informed machine learning (PIML) has attracted considerable attention for its ability to incorporate governing physical principles into the learning framework, with the aim of improving generalization and 
reducing dependence on labeled training data
\cite{karniadakis2021physics,hao2022physics,meng2025physics}. 
However, the practical applicability of PIML is often limited by its sensitivity to variations in problem settings. Achieving consistent predictive performance across problems involving geometric variability, varying spatial and temporal resolution, and nonlinear or history-dependent behavior remains challenging \cite{yuan2022towards,hamdi2026towards,lourencco2025use,fuhg2023modular}. 
In such settings, discrepancies between training and evaluation conditions can lead to physically inconsistent predictions, indicating that the central difficulty lies not only 
in approximation accuracy, but in maintaining the underlying physical structure of the solution across varying problem settings.

\begin{figure}[hbt!]
  \centering
  \includegraphics[width=\linewidth]{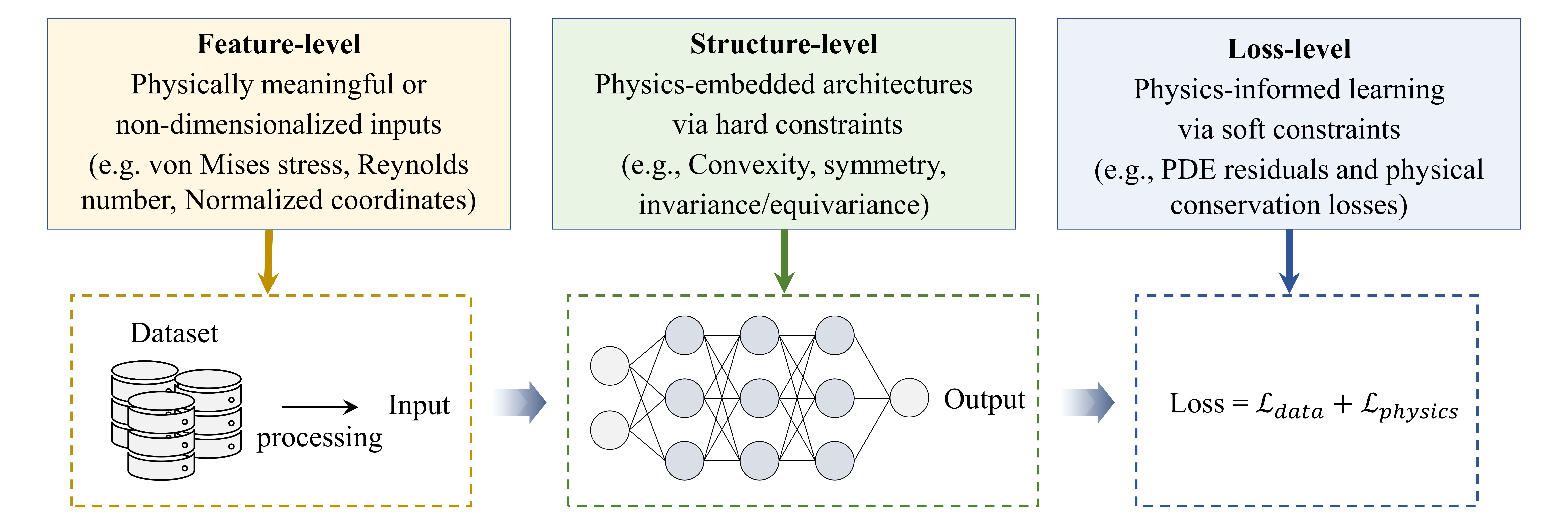}
  \caption{Three levels of physics integration in Physics-Informed Machine Learning (PIML). The classification is based on where the physical principles are incorporated: (i) feature-level via input transformation, (ii) structure-level via neural network architecture design, and (iii) loss-level via residual-based optimization constraints.}
  \label{fig:piml_integration_levels}
\end{figure}

Recent efforts in PIML have therefore focused on how physical knowledge is incorporated into the learning framework. Existing approaches can be broadly categorized into three levels of integration: feature-level, loss-level, and structure-level, as illustrated in Fig.~\ref{fig:piml_integration_levels}   \cite{moon2025physics}. Feature-level strategies introduce physically meaningful or dimensionless input representations, improving learning efficiency and providing invariance with respect to problem scale, but they do not directly constrain the predicted solution \cite{liu2021physics,faegh2025review,leng2025physics}.
Loss-level approaches incorporate governing equations through residual-based loss terms, promoting consistency with the underlying PDE during training. However, these constraints are enforced only implicitly through the optimization process and are not guaranteed to be satisfied in the trained model \cite{cuomo2022scientific,luo2025physics,wang2025kolmogorov}.
Structure-level approaches embed physical priors directly into the model architecture, for example by enforcing symmetry properties, convexity, or invariance and equivariance with respect to coordinate transformations, thereby restricting the admissible function space to physically consistent solutions \cite{thakolkaran2025can,hsu2026accurate,kalina2025physics}. 
Such constraints, however, typically encode general physical properties rather than enforcing the full governing equations, and their applicability is often limited to specific problem classes. 
As a result, each level addresses a distinct aspect of the problem, and no single 
level of integration alone ensures both physical consistency and robustness across varying conditions, suggesting that combining multiple levels of physical knowledge is important for improving model reliability. This limitation becomes particularly pronounced for problems governed by nonlinear and history-dependent PDEs, where the solution evolves incrementally and each state depends on the accumulated response from all prior steps. Ensuring physical consistency throughout this incremental process remains a challenge when data-driven components are applied without explicit coupling to the underlying solution procedure.

To address these limitations, hybrid approaches that couple data-driven surrogate models with conventional numerical solvers have emerged as a promising direction. Rather than replacing physics-based solvers entirely, these approaches embed data-driven components within established computational procedures, restricting their role to selected subproblems while preserving the governing structure of the simulation \cite{mitusch2021hybrid,wang2025shear,BARAL2026107075,pantidis2026integrated,WANG2025109783}. Related efforts in finite element analysis have also demonstrated the potential of replacing selected components of the FEM workflow with learned models, rather than substituting the entire simulation procedure \cite{jung2020deep}. This distinction is particularly important for problems involving nonlinear evolution and history dependence, where the state at each step is determined by the accumulated response from all previous steps. In such settings, the solution is constructed through a sequential update process in which equilibrium and consistency are maintained at each increment.

Among the most computationally demanding examples of such nonlinear, history-dependent systems is phase-field fracture modeling, which has been extensively developed and widely adopted over the past decade due to its ability to represent complex crack phenomena within a unified variational framework. By formulating fracture as the minimization of a coupled energy functional, the phase-field approach enables the simulation of crack initiation, propagation, and branching without the need for explicit crack tracking \cite{miehe2010phase,kuhn2010continuum,ambati2015review,jeong2018phase}. However, this formulation introduces a coupled, nonlinear, and history-dependent PDE system that is solved incrementally. In addition, the diffused representation of a sharp crack requires sufficiently fine spatial and temporal resolution to accurately capture the highly localized damage field. These characteristics lead to substantial computational cost and pose challenges for surrogate modeling, particularly in maintaining stability and consistency throughout the loading history.

In response, machine learning-based surrogates have been developed to accelerate phase-field fracture simulations. These approaches include fully data-driven models that directly approximate the evolution of the phase-field variable \cite{goswami2022physics,perera2023dynamic,kiyani2025predicting,FENG2025110800}, 
as well as physics-informed methods that incorporate governing equations through residual-based constraints \cite{manav2024phase,FENG2025118284,NING2023116430}. 
While these methods can reduce computational cost in specific settings, their applicability remains limited by the need for extensive training data, limited generalization to changes in geometry and loading conditions, and difficulty in capturing the path-dependent nature of crack evolution. In particular, models that approximate the full solution trajectory, including recent spatiotemporal generative approaches for fracture mechanics \cite{park2024deep}, are required to learn spatial, temporal, material, and discretization-dependent behavior within a single mapping. This strategy can increase model complexity and data requirements, especially for nonlinear and history-dependent PDE systems in which the solution evolves through accumulated incremental states. These limitations suggest that the central challenge is not only to increase model capacity or training data, but to identify which components of the governing solution process are most suitable for data-driven approximation. As a result, achieving consistent performance across different configurations requires formulations that preserve the underlying incremental solution structure while selectively introducing data-driven components.

In this work, we propose a hybrid GNN--FEM framework for phase-field fracture modeling, with the objective of achieving physically consistent and generalizable predictions for problems characterized by geometric variability, resolution sensitivity, and history-dependent behavior. The proposed approach integrates a GNN into the conventional staggered solution procedure, where the GNN approximates the phase-field evolution at each load increment while the displacement field is computed using FEM to enforce mechanical equilibrium and boundary conditions. By preserving the incremental structure of the governing equations, the framework maintains consistency with the underlying history-dependent evolution without requiring the surrogate to approximate the full solution trajectory. Physical knowledge is incorporated in a unified manner through dimensionless feature representations, a physics-informed loss based on the governing phase-field equation, and a graph-based architecture operating directly on mesh-based domains.

The main contributions of this work are as follows:
\begin{itemize}
    \item \textbf{GNN-integrated hybrid framework.} A GNN surrogate is introduced within the staggered FEM procedure to approximate the phase-field evolution in a FEM framework for phase-field fracture. The framework selectively replaces the computationally intensive phase-field update while preserving the incremental solution structure and physical consistency by retaining the FEM-based displacement solver throughout the loading history.
    \item \textbf{Multi-level physics integration.} Physical knowledge is incorporated simultaneously through dimensionless feature design, a physics-based residual loss, and a graph-based architecture operating on mesh-based domains, enabling generalization across varying geometries, discretizations, loading conditions, and material properties without retraining.
    \item \textbf{Systematic generalization study.} The framework is evaluated through comprehensive numerical experiments across varying geometries, discretizations, loading conditions, and material properties, providing a systematic benchmark for generalizable surrogate modeling in nonlinear, history-dependent fracture problems.
\end{itemize}
The remainder of this paper is organized as follows. Section~2 introduces the 
phase-field fracture formulation and the staggered solution scheme, and identifies 
the key challenges for generalizable surrogate modeling. Section~3 describes the 
proposed hybrid GNN--FEM framework. Section~4 presents numerical validation results. 
Section~5 provides a discussion of the results, and conclusions are drawn in 
Section~6.

\FloatBarrier

\section{Preliminaries}

\subsection{Crack phase-field model}
\label{subsec:pf_model}

\begin{figure}[hbt!]
\centering
\includegraphics[width=\linewidth]{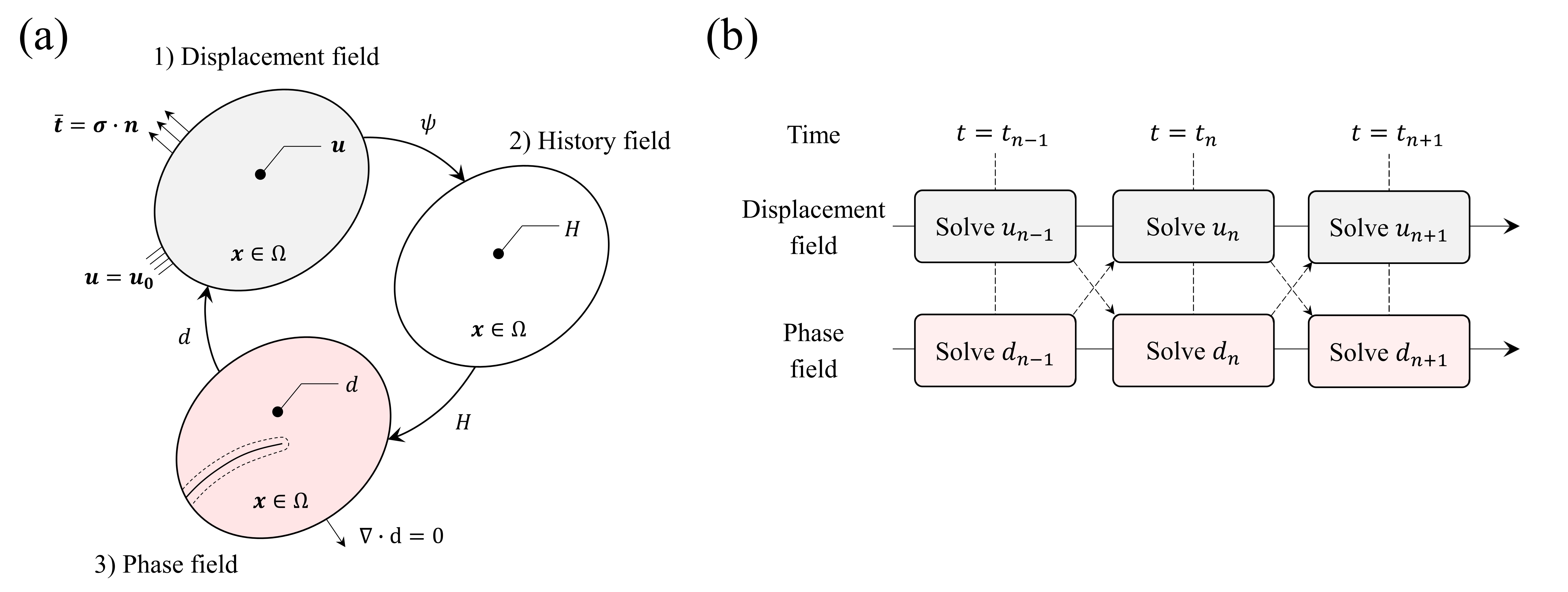}
\caption{Schematic illustration of the phase-field fracture model. 
(a) Coupling between displacement, history, and phase fields. 
(b) Incremental staggered solution scheme, where each field is solved using the converged solution from the previous step.}
\label{fig:pf_intro}
\end{figure}

The phase-field approach to fracture regularizes the sharp crack surface as a diffuse damage zone, enabling crack initiation, propagation, and branching to be captured within a fixed finite element mesh without explicit crack tracking. The governing equations are derived from a variational principle, yielding a coupled system for the displacement field and the phase-field variable that can be discretized and solved using finite element procedures, as illustrated in Fig.~\ref{fig:pf_intro}(a). In the present work, attention is restricted to isotropic, linear elastic solids undergoing brittle fracture under quasi-static loading, and the formulation is summarized below.

Let $\Omega \subset \mathbb{R}^n$ $(n = 2, 3)$ denote the body domain with boundary $\partial\Omega = \Gamma_u \cup \Gamma_t$, where $\Gamma_u$ and $\Gamma_t$ represent the displacement and traction boundaries, respectively, and $\mathbf{n}$ denotes the outward unit normal to $\partial\Omega$. The displacement field is denoted by $\mathbf{u}: \Omega \rightarrow \mathbb{R}^n$, and the crack phase-field variable by $d: \Omega \rightarrow [0,1]$, where $d = 0$ corresponds to the intact state and $d = 1$ to a fully developed crack. 
Following the AT2 formulation \cite{MIEHE20102765}, the crack surface density functional is defined as
\begin{equation}
\gamma(d, \nabla d) = \frac{1}{2\ell_c} d^2 + \frac{\ell_c}{2} |\nabla d|^2,
\end{equation}
where $\ell_c > 0$ is the characteristic length scale parameter controlling the width of the diffuse crack region. The corresponding fracture energy is given by
\begin{equation}
\mathcal{G}(d) = \int_\Omega G_c \, \gamma(d, \nabla d)\, dV,
\end{equation}
where $G_c > 0$ denotes the critical energy release rate.

The total potential energy functional of the system is defined as
\begin{equation}
\Pi(\mathbf{u}, d)
=\int_\Omega g(d)\, \psi_0(\boldsymbol{\varepsilon}(\mathbf{u}))\, dV
+\int_\Omega G_c \, \gamma(d, \nabla d)\, dV
-\int_\Omega \mathbf{b} \cdot \mathbf{u}\, dV
-\int_{\Gamma_t} \bar{\mathbf{t}} \cdot \mathbf{u}\, dS,
\end{equation}
where $\boldsymbol{\varepsilon}(\mathbf{u}) = \tfrac{1}{2}(\nabla \mathbf{u} + \nabla \mathbf{u}^\top)$ 
denotes the infinitesimal strain tensor, $\psi_0(\boldsymbol{\varepsilon})$ is the elastic strain energy density of the undamaged material, and $g(d)$ is the degradation function satisfying $g(0)=1$, $g(1)=0$, and $g'(d)\le 0$, while $\mathbf{b}$ and $\bar{\mathbf{t}}$ denote the body force per unit volume and prescribed traction on $\Gamma_t$, respectively. 
Taking the first variation of $\Pi(\mathbf{u}, d)$ yields the governing equations
\begin{equation}
\nabla \cdot \boldsymbol{\sigma}(\mathbf{u}, d) + \mathbf{b} = \mathbf{0}
\quad \text{in } \Omega,
\end{equation}
\begin{equation}
g'(d)\, H
+ G_c \left( \frac{1}{\ell_c} d - \ell_c \Delta d \right) = 0
\quad \text{in } \Omega,
\end{equation}
subject to the boundary conditions
\begin{equation}
\mathbf{u} = \bar{\mathbf{u}} \quad \text{on } \Gamma_u,
\qquad\boldsymbol{\sigma}(\mathbf{u}, d)\,\mathbf{n} = \bar{\mathbf{t}} \quad \text{on } \Gamma_t,
\end{equation}
\begin{equation}
\nabla d \cdot \mathbf{n} = 0 \quad \text{on } \partial \Omega.
\end{equation}
Here, $H$ denotes the history-dependent crack-driving field, introduced to enforce irreversibility, and is defined in the present work following the hybrid phase-field formulation \cite{ambati2015review, jeong2018phase} as
\begin{equation}
H(\mathbf{x}, t)
=\max_{\tau \in [0,t]} \psi_0^+(\boldsymbol{\varepsilon}(\mathbf{u}(\mathbf{x}, \tau))),
\end{equation}
where $\psi_0^+$ denotes the tensile part of the elastic strain energy density. To evaluate $\psi_0^+$, the strain energy is decomposed into tensile and compressive contributions as
\begin{equation}
\psi_0(\boldsymbol{\varepsilon}) = \psi_0^+(\boldsymbol{\varepsilon}) + \psi_0^-(\boldsymbol{\varepsilon}),
\end{equation}
and, using the spectral decomposition of the strain tensor,
\begin{equation}
\boldsymbol{\varepsilon} = \sum_{a=1}^n \varepsilon_a \, \mathbf{n}_a \otimes \mathbf{n}_a,
\end{equation}
the split energies are expressed as
\begin{equation}
\psi_0^{\pm}(\boldsymbol{\varepsilon})
=\frac{\lambda}{2} \langle \mathrm{tr}(\boldsymbol{\varepsilon}) \rangle_{\pm}^2+\mu \, \mathrm{tr}(\boldsymbol{\varepsilon}_{\pm}^2),
\end{equation}
where $\lambda$ and $\mu$ are the Lam\'{e} constants, and the corresponding undamaged 
stress tensor is defined as
\begin{equation}
\boldsymbol{\sigma}_0 = \frac{\partial \psi_0}{\partial \boldsymbol{\varepsilon}}.
\end{equation}
The stress tensor in the damaged configuration is then given by
\begin{equation}
\boldsymbol{\sigma}(\mathbf{u}, d)
=
g(d)\, \boldsymbol{\sigma}_0(\boldsymbol{\varepsilon}(\mathbf{u})),
\end{equation}
which corresponds to the hybrid phase-field formulation, where crack evolution is governed by the tensile energy while stiffness degradation is applied to the total stress tensor. It is noted that this hybrid phase-field formulation is not strictly consistent with a variational energy minimization principle. However, it is widely adopted in practice because it effectively suppresses unphysical crack growth under compressive loading, which can otherwise arise from the use of purely energy-based degradation models.

Representative forms of the degradation function $g(d)$ are summarized in Table~\ref{tab:gd_variants}. Depending on the choice of $g(d)$, the resulting phase-field evolution equation may become either linear or nonlinear through the form of $g'(d)$ (e.g., a quadratic form leads to a linear equation). In this study, the rational form is deliberately adopted to obtain a nonlinear phase-field equation, providing a more computationally demanding and motivating setting for surrogate-based acceleration. It should be noted, however, that the framework developed in this study is not dependent on a specific choice of $g(d)$ and remains applicable to other degradation models.

\begin{table}[hbt!]
\centering
\small
\caption{Representative degradation functions $g(d)$ used in phase-field fracture models \cite{sargado2018high, livingston2025inference}.}
\label{tab:gd_variants}
\begin{tabular}{p{0.25\linewidth} p{0.35\linewidth}}
\toprule
Type & $g(d)$ \\
\midrule
Quadratic &
$(1 - d)^2$ \\

Cubic &
$(3 - s_0)(1 - d)^2 - (2 + s_0)(1 - d)^3$ \\

Quartic &
$4(1 - d)^3 - 3(1 - d)^4$ \\

Power-law &
$(1 - d)^p$ \\

Rational &
$\displaystyle \frac{(1 - d)^2}{(1 - d)^2 + d}$ \\
\bottomrule
\end{tabular}
\end{table}

\FloatBarrier

\FloatBarrier

\subsection{Staggered solution scheme}
\label{subsec:staggered}

The phase-field fracture problem results in a coupled system of governing equations for the displacement field $\mathbf{u}$ and the phase-field variable $d$. A monolithic solution strategy that simultaneously solves for both fields requires special treatment of the irreversibility constraint $\dot{d} \geq 0$ and often encounters convergence difficulties, particularly near crack initiation and propagation. As a result, a staggered solution scheme is widely adopted, in which the governing equations are decomposed into two subproblems: a displacement problem and a phase-field evolution problem. These are solved sequentially within each load step using the converged solution from the previous step. This approach enables each subproblem to be treated with dedicated iterative procedures while enforcing irreversibility in a straightforward manner, at the cost of a splitting error that diminishes as the load increment decreases.

Let $t_0 < t_1 < \cdots < t_N$ denote the pseudo-time (load) steps, and $(\mathbf{u}_n, d_n, H_n)$ the converged solution at step $t_n$. Given $(\mathbf{u}_{n-1}, d_{n-1}, H_{n-1})$, the objective at step $t_n$ is to determine $(\mathbf{u}_n, d_n)$ such that the governing equations are satisfied together with the irreversibility condition
\begin{equation}
d_n(\mathbf{x}) \geq d_{n-1}(\mathbf{x}),
\qquad 
H_n(\mathbf{x}) = \max\!\Big(H_{n-1}(\mathbf{x}),\;
\psi_0^+(\boldsymbol{\varepsilon}(\mathbf{u}_n(\mathbf{x})))\Big).
\end{equation}
To solve the problem numerically, the domain is discretized in space using finite element approximations, where $\mathcal{T}_h$ denotes a discretization of $\Omega$, and $\mathcal{V}_h \subset [H^1(\Omega)]^n$ and $\mathcal{W}_h \subset H^1(\Omega)$ denote the corresponding function spaces for the displacement and phase-field variables. The weak forms of the two subproblems are given by
\begin{equation}
\int_\Omega \boldsymbol{\sigma}(\mathbf{u}, d) : \boldsymbol{\varepsilon}(\mathbf{v})\, dV
=
\int_\Omega \mathbf{b} \cdot \mathbf{v}\, dV
+
\int_{\Gamma_t} \bar{\mathbf{t}} \cdot \mathbf{v}\, dS,
\qquad \forall\, \mathbf{v} \in \mathcal{V}_h,
\end{equation}
\begin{equation}
\int_\Omega \Big(
g'(d)\, H\, w
+ G_c \big( \tfrac{1}{\ell_c}\, d\, w + \ell_c\, \nabla d \cdot \nabla w \big)
\Big)\, dV = 0,
\qquad \forall\, w \in \mathcal{W}_h,
\end{equation}
for admissible test functions $\mathbf{v}$ and $w$, which lead to the residual equations
\begin{equation}
\mathbf{R}_u(\mathbf{u}_h, d_h) = \mathbf{0},
\qquad
\mathbf{R}_d(d_h, H_h) = \mathbf{0}.
\end{equation}

At each load step $t_n$, the displacement field $\mathbf{u}_n$ is first obtained by solving
\begin{equation}
\mathbf{R}_u(\mathbf{u}_n, d_{n-1}) = \mathbf{0},
\end{equation}
with $d_{n-1}$ held fixed, followed by the phase-field update
\begin{equation}
\mathbf{R}_d(d_n, H_{n-1}) = \mathbf{0},
\end{equation}
with $H_{n-1}$ fixed. In contrast to a monolithic approach that solves for $\mathbf{u}$ and $d$ simultaneously, the staggered scheme treats the two subproblems sequentially, each solved using Newton--Raphson iteration, thereby simplifying the solution procedure. The irreversibility constraint is enforced through $d_n \geq d_{n-1}$, and the history field is updated as
\begin{equation}
H_n(\mathbf{x})
=\max\!\Big(H_{n-1}(\mathbf{x}),\; \psi_0^+(\boldsymbol{\varepsilon}(\mathbf{u}_n(\mathbf{x})))\Big).
\end{equation}
The structure of the staggered scheme is summarized in Algorithm~\ref{alg:staggered} and illustrated in Fig.~\ref{fig:pf_intro}(b).

\begin{algorithm}[hbt!]
\caption{Incremental staggered scheme for phase-field fracture}
\label{alg:staggered}
\begin{algorithmic}[1]
\State \textbf{Input:} finite element mesh $\mathcal{T}_h$, load steps $\{t_n\}_{n=0}^N$
\For{$n=1$ to $N$}
    \State Apply loading and boundary conditions at step $t_n$
    \State Solve $\mathbf{R}_u(\mathbf{u}_n,d_{n-1})=\mathbf{0}$ by Newton--Raphson iteration
    \State Solve $\mathbf{R}_d(d_n,H_{n-1})=\mathbf{0}$ by Newton--Raphson iteration
    \State Update $H_n \gets \max\!\big(H_{n-1},\psi_0^+(\boldsymbol{\varepsilon}(\mathbf{u}_n))\big)$
    \State Enforce irreversibility: $d_n \gets \max(d_n,d_{n-1})$
\EndFor
\end{algorithmic}
\end{algorithm}

\FloatBarrier

\subsection{Challenges toward generalizable surrogate modeling}
\label{subsec:challenges}
Constructing a surrogate model that is both computationally efficient and broadly generalizable remains a fundamental challenge in SciML. Although phase-field fracture serves as the primary application context in this work, the difficulties discussed here are not specific to this problem. Rather, they arise more generally when data-driven approximations are applied to nonlinear, history-dependent PDE systems. The key challenges are summarized below.
\setlength{\parindent}{0pt}
\setlength{\parskip}{6pt}

\textbf{\underline{Nonlinearity and history dependence.}}
Many physical systems are governed by equations that are both strongly nonlinear and path-dependent. In such cases, the current state cannot be determined from instantaneous inputs alone but depends on the entire loading history. Direct mappings from input parameters to solution fields therefore fail to reproduce correct responses under non-trivial loading sequences. This issue is particularly pronounced in fracture problems, where crack evolution is irreversible and governed by the historical maximum of the elastic energy. Effective surrogate models must operate within an incremental framework that explicitly propagates relevant history variables across load steps.

\textbf{\underline{Geometric variability and domain topology.}}
Surrogate models are often expected to generalize across different geometries, yet grid-based architectures are inherently tied to fixed domain representations and cannot be directly transferred to new geometries. This limitation becomes critical when domains exhibit irregular boundaries, internal interfaces, or localized features, as commonly encountered in fracture mechanics. Addressing such variability requires representations that encode the connectivity and spatial relationships of the computational domain without assuming a fixed underlying structure.

\textbf{\underline{Sensitivity to discretization.}}
The accuracy of nonlinear PDE solutions depends strongly on discretization choices, including mesh resolution and load increment size. Models trained at a specific resolution generally do not transfer to different discretizations. In phase-field fracture, this issue is further exacerbated by the requirement that the element size be tied to the regularization length scale, leading to varying meshes across problem instances. Consequently, surrogate models are required to process solution and geometric information consistently across discretizations, rather than relying on fixed input structures.

\textbf{\underline{Variability in loading and boundary conditions.}}
The solution of boundary value problems is highly sensitive to loading paths and boundary conditions. Training on limited scenarios often results in poor generalization to unseen conditions, particularly when loading induces complex, spatially varying stress states. Purely data-driven models lack mechanisms to enforce physical consistency under such variations, making reliable generalization across diverse loading and boundary conditions inherently challenging.

\textbf{\underline{Variability in material properties.}}
Variations in material properties directly influence the relationship between loading and solution fields, yet models trained on a fixed parameter set typically do not generalize to other parameter regimes when inputs are expressed in dimensional form, as dimensional features entangle the effects of material properties, geometry, and loading. This entanglement makes it difficult to isolate transferable physical relationships across varying material conditions.

\textbf{\underline{Limited availability of training data.}}
Generating training data for nonlinear PDE systems is computationally expensive, as each sample requires a full numerical solve. As a result, available datasets are often limited in size and coverage. Under such constraints, purely data-driven models are prone to overfitting and exhibit poor generalization. While the governing equations impose strong constraints on admissible solutions, this structure is not exploited in standard supervised learning. Incorporating physical constraints into the training process therefore provides an effective form of regularization.
\setlength{\parindent}{15pt}
\setlength{\parskip}{0pt}

These challenges highlight the inherent difficulty in developing surrogate models that are both efficient and broadly generalizable for nonlinear, history-dependent PDE systems. Addressing these issues requires formulations that account for incremental solution structure, operate on flexible domain representations, and incorporate physically meaningful features and constraints. These considerations naturally point toward hybrid modeling strategies that combine data-driven approaches with physics-based solvers, as explored in the following section.

\section{Proposed Hybrid GNN--FEM Framework}

\subsection{Overview of the hybrid formulation}
\label{subsec:overview}

The proposed hybrid GNN--FEM framework embeds a graph neural network (GNN) surrogate within the conventional staggered solution scheme described in Section~\ref{subsec:staggered}. The surrogate is constructed by incorporating physical knowledge through feature design, loss formulation, and network architecture, enabling it to operate consistently with the governing equations. This design directly addresses the challenges outlined in Section~\ref{subsec:challenges}, particularly those arising from history dependence, geometric variability, and sensitivity to discretization. Rather than approximating the full solution mapping, the framework preserves the incremental structure of the governing equations and introduces data-driven components only in parts of the solution process that are more amenable to approximation. As a result, the model operates within the physical solution procedure while maintaining consistency across load steps.

\begin{figure}[hbt!]
\centering
\includegraphics[width=\linewidth]{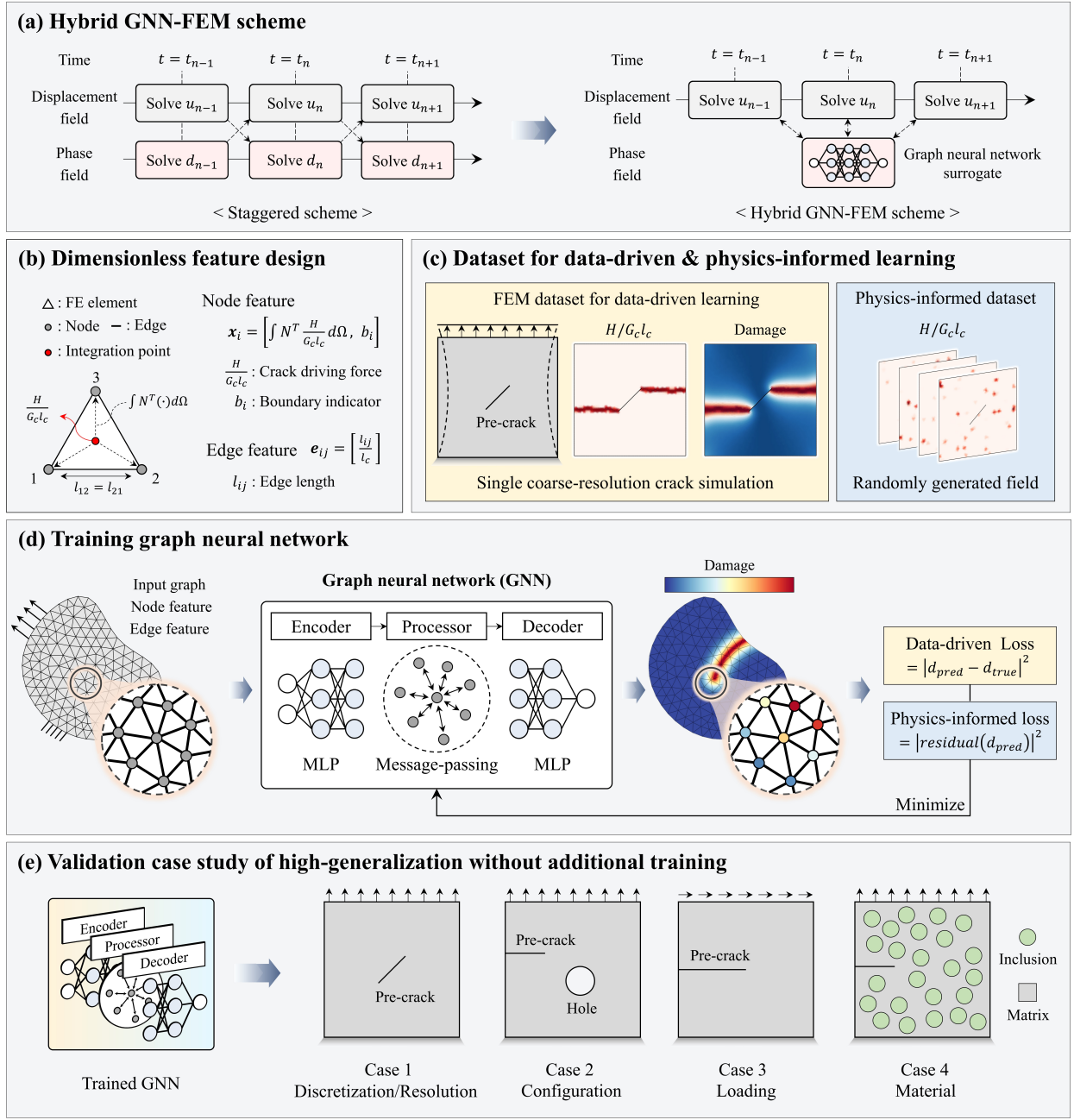}
\caption{Overview of the proposed hybrid GNN--FEM framework. 
(a) Comparison between the conventional staggered scheme and the hybrid GNN--FEM scheme, 
where the phase-field solve is replaced by the GNN surrogate. 
(b) Graph representation and dimensionless feature design. 
(c) Composition of the training dataset. 
(d) GNN encoder--processor--decoder architecture and corresponding training loss. 
(e) Validation cases used to assess generalization performance.}
\label{fig:framework}
\end{figure}

Specifically, the FEM-based displacement solver is retained, while the phase-field update is replaced by the GNN surrogate, as illustrated in Fig.~\ref{fig:framework}(a). This decomposition reflects the distinct characteristics of the two subproblems with respect to surrogate modeling. The displacement subproblem is governed by equilibrium equations with prescribed loading and boundary conditions, and its accurate prediction requires enforcing global mechanical consistency across varying configurations, which remains difficult for data-driven models. In contrast, the phase-field subproblem is a scalar evolution equation driven by local quantities and is therefore more suitable for surrogate approximation. At each load step, the displacement and history fields are updated using FEM, after which the GNN surrogate predicts the phase-field from a graph-based description of the domain and the history field, with irreversibility enforced as a post-processing step. Through this design, the framework enables surrogate modeling within an incremental solution process while maintaining robustness across different geometries, discretizations, and loading conditions, without requiring direct approximation of the full solution trajectory.

\FloatBarrier

\subsection{Graph representation and feature design}
\label{subsec:graph}

The finite element mesh is represented as a graph 
$\mathcal{G} = (\mathcal{V}, \mathcal{E})$, where the node set $\mathcal{V}$ 
corresponds to the finite element nodes and the edge set $\mathcal{E}$ encodes the connectivity between adjacent nodes. An edge $(i, j) \in \mathcal{E}$ is introduced for each pair of nodes sharing a finite element, so that the graph topology is directly inherited from the mesh connectivity.

Each node $i \in \mathcal{V}$ is assigned a feature vector
\begin{equation}
\mathbf{x}_i = \left[ \left(\int N^T \frac{H}{G_c \ell_c}\, d\Omega\right)_i,\; b_i \right],
\end{equation}
where $N$ denotes the finite element shape function matrix. The first feature is the normalized history field projected onto nodal values. The history field $H$, originally defined at integration points, is first normalized by $G_c \ell_c$ to obtain a dimensionless quantity, and subsequently projected onto nodal values through the shape functions, i.e., by applying the standard finite element projection operator $\int N^T (\cdot)\, d\Omega$. This operation reuses the projection procedure already available in the displacement solve and introduces no additional computational cost. The second feature is a binary boundary indicator $b_i$, which takes the value one if node $i$ lies on any boundary of the domain, 
including pre-crack surfaces, and zero otherwise. While this normalization is defined for two-dimensional problems, it extends naturally to three dimensions by replacing $G_c \ell_c$ with $G_c \ell_c^2$.

Each edge $(i, j) \in \mathcal{E}$ is assigned a feature
\begin{equation}
\mathbf{e}_{ij} = \left[ \frac{l_{ij}}{\ell_c} \right],
\end{equation}
where $l_{ij}$ denotes the length of edge $(i, j)$. Normalizing the edge length by the characteristic length scale ensures a consistent representation across different discretization levels. In phase-field fracture simulations, the mesh size $h$ is typically chosen in relation to the regularization length scale $\ell_c$, such that a fixed ratio $h/\ell_c$ is maintained to properly resolve the diffuse crack profile, commonly with $h \approx \ell_c/2$. The dimensionless form $l_{ij}/\ell_c$ therefore allows the feature representation to remain consistent across problems with different absolute length scales, as long as the ratio between mesh size and regularization length scale is preserved.

The use of dimensionless node and edge features is central to the generalization capability of the proposed framework. Normalizing the history field by $G_c \ell_c$ ensures that problems with different 
material properties but identical physical states yield the same dimensionless 
feature values, reducing explicit dependence on material-property scales and 
allowing a single trained model to be applied across different values of $G_c$ 
and the elastic constants. Similarly, normalizing edge lengths by $\ell_c$ removes explicit dependence on discretization, enabling consistent predictions across meshes of varying resolution. The node and edge features are illustrated schematically in Fig.~\ref{fig:framework}(b).

\subsection{Graph neural network as a surrogate for phase-field evolution}
\label{subsec:gnn}

Among the various neural network architectures, a GNN is adopted as the surrogate model for the phase-field evolution due to its ability to operate on unstructured mesh representations and to capture local interactions between neighboring nodes, which are intrinsic to the governing phase-field equation. The surrogate adopts an encoder–processor–decoder architecture, in which graph-based input features are mapped to latent variables, updated through message passing, and decoded to obtain the phase-field value at each node.

\textbf{Encoder.} The input to the model consists of node features 
$\mathbf{x}_i \in \mathbb{R}^2$ and edge features $\mathbf{e}_{ij} \in \mathbb{R}^1$, 
as defined in Section~\ref{subsec:graph}, and the output is a scalar phase-field 
prediction $\hat{d}_i$ at each node. The first node feature, corresponding to the 
normalized history field, is transformed using a logarithmic mapping of the form 
$\log_{10}(1 + (\cdot))$ prior to encoding, which reduces the dynamic range of the 
input and improves numerical stability during training. The node and edge features 
are then independently encoded into a latent space of dimension $H_\mathrm{lat}$ 
through separate MLPs, followed by layer normalization:
\begin{equation}
\mathbf{z}_i^{(0)} = \mathrm{LN}\!\left(\mathrm{MLP}_\mathrm{node}(\mathbf{x}_i)\right),
\qquad
\mathbf{f}_{ij}^{(0)} = \mathrm{LN}\!\left(\mathrm{MLP}_\mathrm{edge}(\mathbf{e}_{ij})\right),
\end{equation}
where $\mathbf{z}_i^{(0)} \in \mathbb{R}^{H_\mathrm{lat}}$ and 
$\mathbf{f}_{ij}^{(0)} \in \mathbb{R}^{H_\mathrm{lat}}$ denote the initial latent 
node and edge variables, respectively. The input follows the dimensionless feature design introduced in Section~\ref{subsec:graph}, allowing the relevant physical quantities to be expressed in a scale-normalized form.

\textbf{Processor.} The initial latent variables $\mathbf{z}_i^{(0)}$ and 
$\mathbf{f}_{ij}^{(0)}$ are subsequently refined through $K$ message-passing 
iterations, yielding updated latent variables $\mathbf{z}_i^{(k)}$ and 
$\mathbf{f}_{ij}^{(k)}$ at each iteration $k = 1, \ldots, K$. At each iteration, 
the edge features are first updated via a residual connection:
\begin{equation}
\mathbf{f}_{ij}^{(k)} = \mathbf{f}_{ij}^{(k-1)} + \mathrm{MLP}_\mathrm{edge}^{(k)}\!\left(
\mathbf{z}_i^{(k-1)} + \mathbf{z}_j^{(k-1)},\;
\left|\mathbf{z}_i^{(k-1)} - \mathbf{z}_j^{(k-1)}\right|,\;
\mathbf{f}_{ij}^{(k-1)}
\right).
\end{equation}
The input is constructed from the sum and absolute difference of the two incident 
node variables, ensuring symmetry with respect to node ordering. This design is 
consistent with the underlying finite element formulation, in which the stiffness 
matrix is symmetric and the interaction between adjacent nodes is reciprocal.
Messages are then computed with pre-layer normalization,
\begin{equation}
\mathbf{m}_{ij}^{(k)} = \mathrm{MLP}_\mathrm{msg}^{(k)}\!\left(
\mathrm{LN}\!\left(
\mathbf{z}_i^{(k-1)} + \mathbf{z}_j^{(k-1)},\;
\left|\mathbf{z}_i^{(k-1)} - \mathbf{z}_j^{(k-1)}\right|,\;
\mathbf{f}_{ij}^{(k)}
\right)\right),
\end{equation}
and aggregated over neighboring nodes by mean pooling:
\begin{equation}
\bar{\mathbf{m}}_i^{(k)} = \frac{1}{|\mathcal{N}(i)|} \sum_{j \in \mathcal{N}(i)} 
\mathbf{m}_{ij}^{(k)},
\end{equation}
where $\mathcal{N}(i)$ denotes the set of nodes adjacent to node $i$. 
The node variables are then updated through a residual connection:
\begin{equation}
\mathbf{z}_i^{(k)} = \mathbf{z}_i^{(k-1)} + 
\mathrm{MLP}_\mathrm{upd}^{(k)}\!\left(
\mathrm{LN}\!\left(\mathbf{z}_i^{(k-1)},\; \bar{\mathbf{m}}_i^{(k)}\right)\right).
\end{equation}
This message-passing procedure propagates information locally through the mesh 
connectivity, enabling the model to operate on domains of arbitrary geometry 
and topology.

\textbf{Decoder.} The final node variables are passed through a decoder MLP to produce the predicted phase-field value at each node:
\begin{equation}
\hat{d}_i = \mathrm{MLP}_\mathrm{dec}\!\left(\mathbf{z}_i^{(K)}\right).
\end{equation}

Since all input features are scalar quantities, the GNN model is invariant under rotations of the coordinate frame, consistent with the isotropic nature of the governing equations. Consequently, a model trained on crack propagation in one direction can generalize to predict crack evolution in other directions without requiring additional training data. This property, however, relies on the assumption of isotropy; in the presence of anisotropic material behavior with direction-dependent responses, an equivariant GNN architecture would be required to properly account for such dependencies. Addressing this extension is beyond the scope of the present work and is left for future work.

\subsection{Physics-informed and data-driven training strategy}
\label{subsec:training}

The training strategy consists of supervised data-driven learning and physics-informed learning, which are trained jointly within a single optimization process. While the two learning processes are based on different datasets, they are integrated through a unified loss function during training. Data-driven learning directly utilizes FEM solutions, whereas physics-informed learning incorporates information from the governing equations, thereby extending the coverage of the training distribution. This strategy is motivated by the limitations of purely data-driven approaches in nonlinear, history-dependent PDE systems, as discussed in Section~\ref{subsec:challenges}, where the availability of high-fidelity simulation data is inherently limited. As a result, the model learns not only from available data but also from the underlying physical laws.

\begin{figure}[hbt!]
\centering
\includegraphics[width=\linewidth]{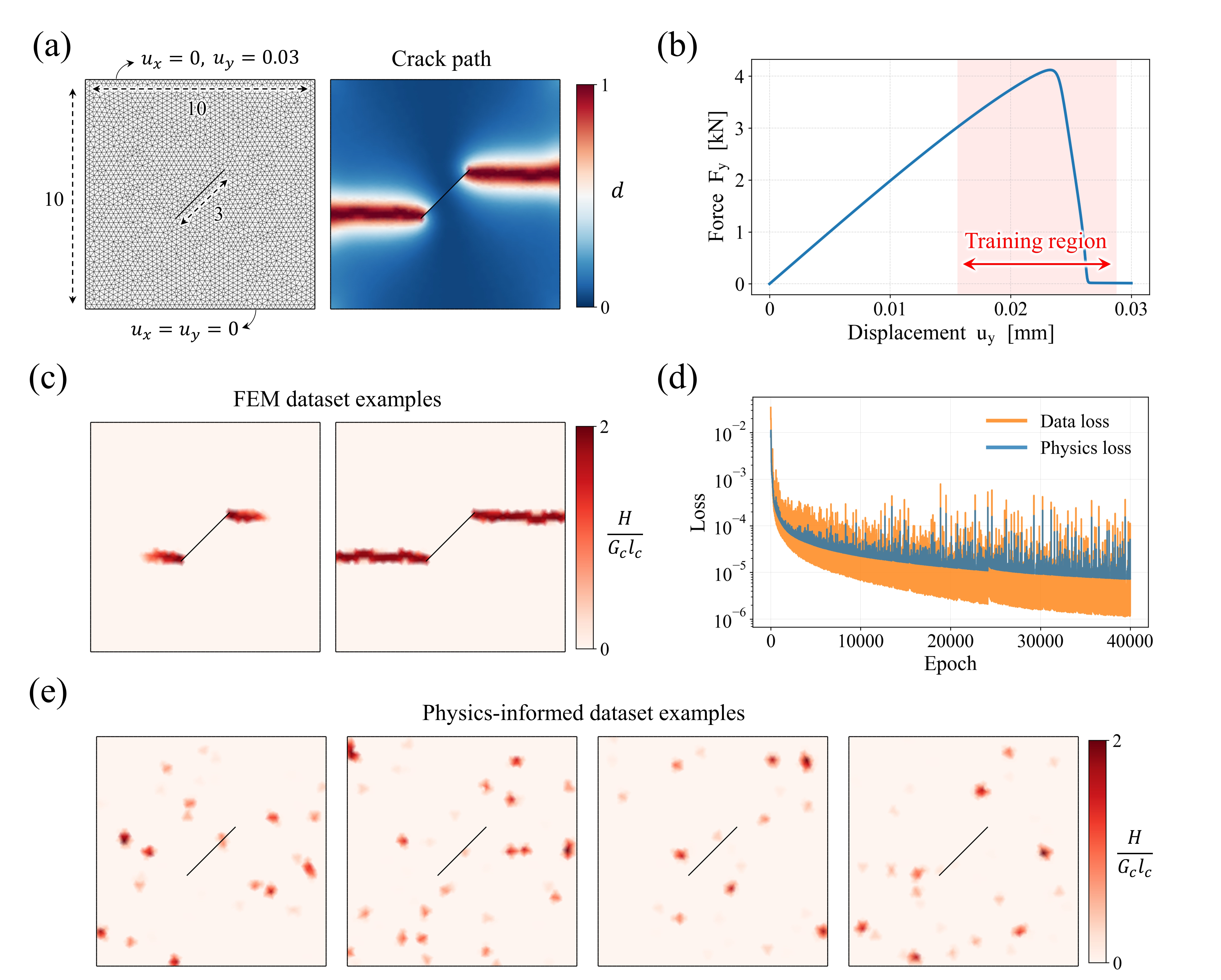}
\caption{Training setup for the GNN surrogate. 
(a) Geometry, loading and boundary conditions, and the resulting crack path of the training simulation. 
(b) Force--displacement response with the selected training region highlighted. 
(c) Representative examples of the normalized history field distribution from the data-driven dataset at load increments 420 and 450. 
(d) Evolution of the data-driven and physics-informed loss components during training. 
(e) Representative examples of the randomly generated normalized history fields used for physics-informed training.}
\label{fig:gnn_training}
\end{figure}

Constructing an FEM dataset for supervised data-driven learning that adequately 
captures the full range of crack evolution scenarios is inherently challenging, 
as it would require a large number of high-fidelity FEM simulations. To mitigate this limitation, a single phase-field fracture simulation is employed in the present work to generate training data, based on the geometry and loading conditions shown in Fig.~\ref{fig:gnn_training}(a), consisting of a square domain with a pre-crack subjected to vertical displacement loading. The material is assumed to be linear elastic with a Young’s modulus $E = 210\mathrm{GPa}$, Poisson’s ratio $\nu = 0.3$, and critical energy release rate $G_c = 2.7\mathrm{N/mm}$. The simulation is performed using a mesh size of $h = 0.2\mathrm{mm}$, a regularization length scale $\ell_c = 0.4\mathrm{mm}$, and 500 load increments. As shown in Fig.~\ref{fig:gnn_training}(a), the relatively large ratio of $\ell_c$ to the domain size produces a diffuse crack profile. This setting is intentionally adopted to enable efficient data generation, as the diffuse crack representation allows the use of a relatively coarse mesh and reduces the computational cost of each simulation while preserving the essential characteristics of crack propagation. Training samples are extracted from 200 increments within the crack-dominant regime, corresponding to increments 251 through 450 in Fig.~\ref{fig:gnn_training}(b), where significant crack evolution occurs. The nodal values of the history field and phase-field, obtained through projection using finite element shape functions, define the input–output pairs for supervised training and are used to compute the data-driven loss. The input features consist of the normalized history field $H/(G_c \ell_c)$ together with the boundary indicator and edge features described in Section~\ref{subsec:graph}. Representative examples of the normalized history field in the dataset are shown in Fig.~\ref{fig:gnn_training}(c).

For the physics-informed learning, a dataset of spatially correlated random fields of $H/(G_c \ell_c)$ is generated, as illustrated in Fig.~\ref{fig:gnn_training}(e). These fields do not represent physically realizable crack states; instead, they are randomly generated to provide diverse history field distributions and broaden the range of input conditions seen during training. Since the corresponding phase-field solutions are not available for such inputs, the GNN is trained using the residual of the governing phase-field equation, which defines the physics-informed loss. In this manner, the model is encouraged to satisfy the governing equations even for 
input states that are not observed in the FEM dataset used for supervised 
data-driven learning. A total of 200 random fields are generated, matching the size of the FEM dataset, and the generation procedure is described in Appendix~A. It is worth noting that the cost of generating the random fields is negligible, and the additional cost introduced during training remains minor compared to that required for generating additional FEM-based datasets.

Consequently, the GNN is trained by minimizing the total loss defined as
\begin{equation}
\mathcal{L} = \mathcal{L}_\mathrm{data} + \lambda_\mathrm{phys}\,\mathcal{L}_\mathrm{phys},
\end{equation}
where $\lambda_\mathrm{phys} \geq 0$ is a weighting coefficient. The data-driven 
loss $\mathcal{L}_\mathrm{data}$ is defined as the mean squared error between the 
predicted and FEM-computed phase-field values,
\begin{equation}
\mathcal{L}_\mathrm{data} = \frac{1}{|\mathcal{V}|}\sum_{i \in \mathcal{V}} 
\left( \hat{d}_i - d_i^\mathrm{FEM} \right)^2,
\end{equation}
while the physics-informed loss $\mathcal{L}_\mathrm{phys}$ is given by the mean 
squared residual of the discretized weak form of the phase-field evolution equation,
\begin{equation}
\mathcal{L}_\mathrm{phys} = \frac{1}{|\mathcal{V}|} \left\| \mathbf{R}_d(\hat{\mathbf{d}}, \mathbf{H}) \right\|^2,
\end{equation}
where $\mathbf{R}_d$ denotes the residual introduced in 
Section~\ref{subsec:staggered}. In the present work, $\lambda_\mathrm{phys} = 1$, 
so that both loss terms contribute at comparable magnitude. The evolution of both 
loss components during training is shown in Fig.~\ref{fig:gnn_training}(d). The hyperparameters related to the GNN architecture and training are summarized in 
Table~\ref{tab:gnn_hyperparams}. 

The trained GNN model is evaluated across a range of validation cases in Section~\ref{sec:validation}. In this process, a comparison is also made between the model trained using both data-driven and physics-informed learning and a counterpart trained using data-driven learning alone, in order to assess the effect of incorporating the physics loss.

\FloatBarrier

\begin{table}[hbt!]
\centering
\caption{Hyperparameters of the GNN architecture and training. 
The same hidden dimension and number of hidden layers are used for all MLP modules in the network.}
\label{tab:gnn_hyperparams}
\begin{tabular}{lc}
\toprule
Parameter & Value \\
\midrule
Latent dimension, $H_{\mathrm{lat}}$          & 32        \\
Number of hidden layers per MLP, $n_l$        & 3         \\
Message-passing iterations, $K$               & 10        \\
Input node feature dimension                 & 2         \\
Input edge feature dimension                 & 1         \\
Output feature dimension                     & 1         \\
Activation function                          & ReLU      \\
\midrule
Optimizer                                    & Adam      \\
Learning rate                                & $10^{-4}$ \\
Training epochs                              & 40{,}000  \\
\midrule
Number of data-driven samples                & 200       \\
Number of physics-informed samples           & 200       \\
Physics loss weight, $\lambda_{\mathrm{phys}}$ & 1         \\
\bottomrule
\end{tabular}
\end{table}

\FloatBarrier


\section{Numerical Validation}
\label{sec:validation}

In this section, the predictive performance and generalization capability of the GNN surrogate are systematically evaluated. The objective of this validation study is to examine whether the model can accurately predict phase-field evolution beyond the training setting while remaining consistent with the governing physics. To this end, the model is assessed under variations in mesh resolution, geometry, loading conditions, and material properties. These variations respectively modify the discretization, domain geometry, boundary conditions, and material parameters, thereby providing a comprehensive evaluation of the robustness and transferability of the proposed hybrid GNN--FEM framework. The GNN surrogate, trained on a single setting as described in Section~\ref{subsec:training}, is applied to all validation cases without any additional training or fine-tuning, so that the results directly reflect its performance across previously unseen scenarios.

For clarity, the reference solutions obtained using the staggered FEM are denoted as \textit{Full FEM}, while the proposed approach is referred to as \textit{Hybrid GNN--FEM}. Models trained using both data-driven and physics-informed learning are labeled as (PI-DD), and those trained using data-driven learning only are labeled as (DD). All validation simulations are performed using 1000 load increments, which is finer than the temporal discretization used during training, in order to assess the robustness of the model under refined loading paths. Unless otherwise specified, all simulations are conducted using a mesh size of $h = 0.05$, with the ratio $\ell_c = 2h$ maintained to ensure consistent phase-field resolution, under two-dimensional plane stress conditions, and with material properties identical to those used in the training. The computational domain is discretized using triangular finite elements, enabling consistent meshing of geometries with arbitrary shapes and local features such as holes or inclusions, and reflecting the intended application to unstructured domains. The detailed geometric configurations for all validation cases are provided in Appendix~B.

\subsection{Generalization to unseen discretization resolutions}

\begin{figure}[h]
\centering
\includegraphics[width=\linewidth]{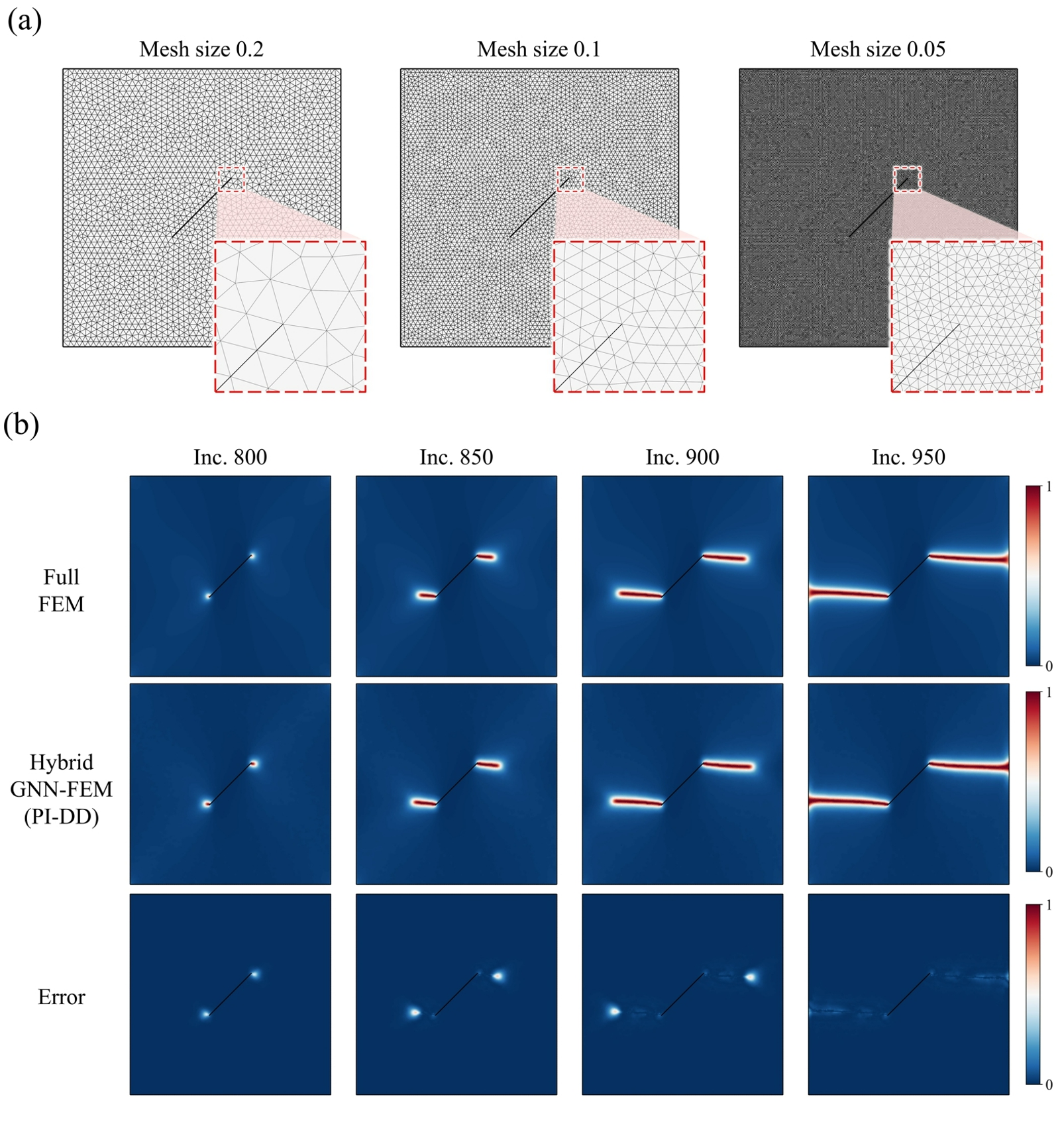}
\caption{Generalization to unseen discretization resolutions. 
(a) Finite element meshes with different resolutions ($h = 0.2$, $0.1$, and $0.05$). 
(b) Comparison of phase-field evolution at selected load increments for Full FEM, Hybrid GNN--FEM (PI-DD), and the corresponding error, shown for the finest discretization ($h = 0.05$).}
\label{fig:val_mesh_field}
\end{figure}

To evaluate the sensitivity of the proposed framework to spatial and temporal discretization, the GNN surrogate is applied to the same problem setting used during training, while varying only the mesh resolution. Simulations are performed with mesh sizes of $h = 0.2$, $0.1$, and $0.05$, where $h = 0.2$ corresponds to the training case, and the corresponding finite element meshes are shown in Fig.~\ref{fig:val_mesh_field}(a). The ratio between the regularization length scale and mesh size is consistently maintained as $\ell_c = 2h$, ensuring that the phase-field remains properly resolved across all spatial discretizations. In addition, all validation simulations are performed using 1000 load increments, which is finer than the temporal discretization used during training. This setup enables assessment of the robustness of the model with respect to both spatial and temporal resolution.

The phase-field evolution predicted by the Hybrid GNN--FEM (PI-DD) is compared with the Full FEM solution in Fig.~\ref{fig:val_mesh_field}(b) for the finest discretization ($h = 0.05$). The Hybrid GNN--FEM (PI-DD) accurately reproduces the crack initiation and propagation behavior observed in the Full FEM solution. The predicted crack path and damage distribution remain consistent, with discrepancies confined to narrow regions near the crack front. This indicates that the model captures the local fracture response without being tied to a specific mesh resolution.

\begin{figure}[h]
\centering
\includegraphics[width=\linewidth]{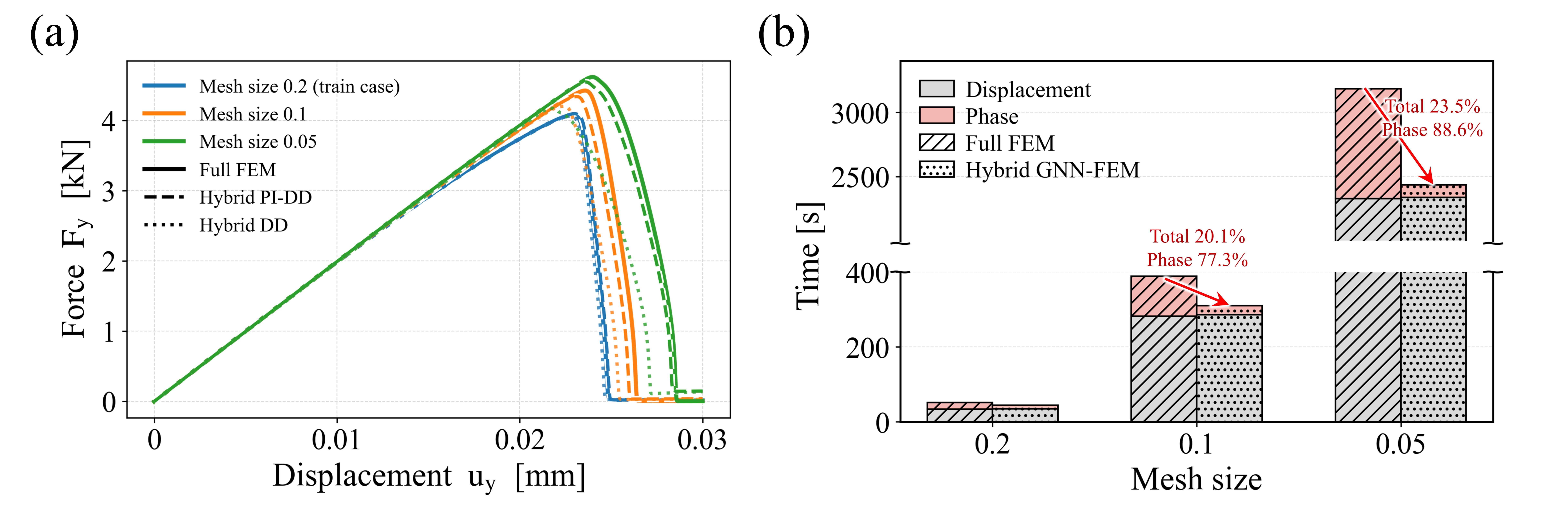}
\caption{Quantitative comparison across discretization resolutions. 
(a) Force--displacement responses for different mesh sizes, comparing Full FEM, Hybrid GNN--FEM (PI-DD), and Hybrid DD. 
(b) Computational cost comparison between Full FEM and Hybrid GNN--FEM, with decomposition into displacement and phase-field contributions.}
\label{fig:val_mesh_Rd_curve}
\end{figure}

The influence of discretization on the force--displacement response is examined in Fig.~\ref{fig:val_mesh_Rd_curve}(a). For the training discretization ($h = 0.2$), both Hybrid GNN--FEM (PI-DD) and Hybrid DD closely match the Full FEM solution. However, as the mesh is refined ($h = 0.1$ and $0.05$), a clear discrepancy in the response becomes evident. The Hybrid DD model progressively deviates from the Full FEM solution, particularly in the post-peak regime, indicating reduced predictive accuracy outside the training discretization. In contrast, the Hybrid GNN--FEM (PI-DD) maintains close agreement with the Full FEM curves across all mesh resolutions. This behavior indicates that physics-informed learning plays a critical role in preserving consistency with the governing equations under discretization changes, allowing the model to retain accuracy even when the discretization differs from that used during training.

The computational cost is compared in Fig.~\ref{fig:val_mesh_Rd_curve}(b). As the mesh is refined, the cost of the Full FEM solution increases significantly due to the higher number of degrees of freedom and the associated nonlinear iterations required to solve the phase-field equation. The Hybrid GNN--FEM (PI-DD) also exhibits an increase in computational cost with mesh refinement, but with a consistently lower overall runtime due to the replacement of the phase-field solve by the GNN surrogate. This avoids the need to iteratively solve the nonlinear phase-field equation at each load step, leading to a reduction in overall simulation time. The total computational time is reduced by approximately 20.1\% for $h = 0.1$ and 23.5\% for $h = 0.05$, with a more pronounced reduction observed in the phase-field component (77.3\% and 88.6\%, respectively). The reduction in computational cost primarily originates from the replacement of the phase-field solve, while the displacement solve continues to dominate the overall computational expense.

The results confirm that the proposed framework maintains predictive accuracy under variations in both spatial and temporal discretization beyond the training configuration, while also providing a consistent reduction in computational cost. 

\FloatBarrier

\FloatBarrier

\subsection{Generalization to unseen geometric configurations}

\begin{figure}[hbt!]
\centering
\includegraphics[width=\linewidth]{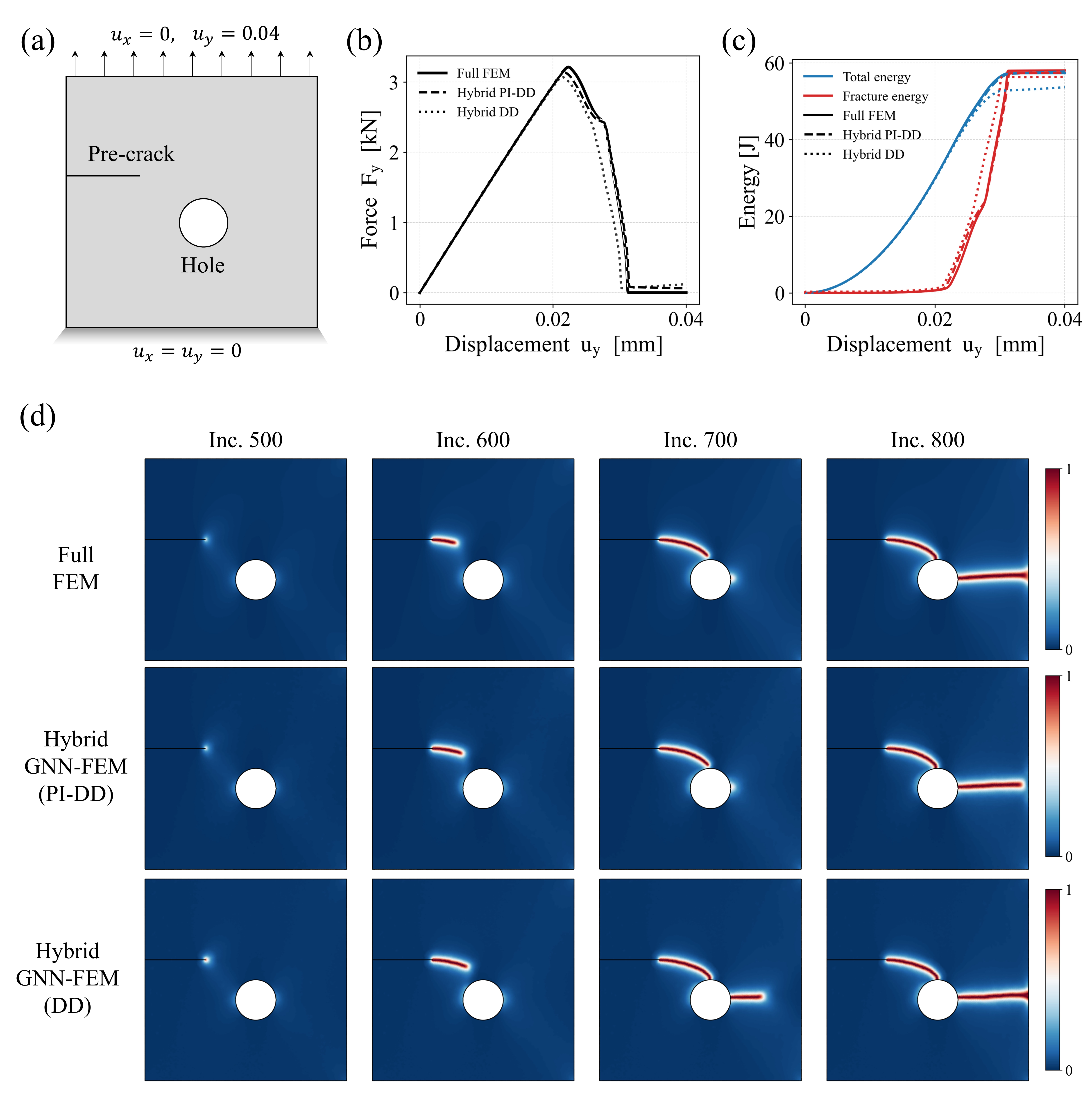}
\caption{Generalization to unseen geometric configurations. 
(a) Problem setup with a modified geometry including a pre-crack and a circular hole. 
(b) Force--displacement responses comparing Full FEM, Hybrid GNN--FEM (PI-DD), and Hybrid DD. 
(c) Evolution of total and fracture energies. 
(d) Phase-field distributions at selected load increments for Full FEM, Hybrid GNN--FEM (PI-DD) and Hybrid DD.}
\label{fig:val_geometry_field}
\end{figure}

Generalization across geometries is evaluated by applying the trained GNN surrogate to a different geometry, as illustrated in Fig.~\ref{fig:val_geometry_field}(a), using a mesh size of $h = 0.05$ and 1000 load increments. A circular hole is introduced in addition to the pre-existing crack, creating a crack--hole interaction that modifies the stress field and introduces additional crack driving effects due to local stress concentration.

The force--displacement behavior and energy evolution are presented in Fig.~\ref{fig:val_geometry_field}(b) and (c). The Hybrid GNN--FEM (PI-DD) closely follows the Full FEM response, capturing both the peak load and the subsequent softening behavior. Consistent agreement is also observed in the energy evolution, where the PI-DD model reproduces the accumulation of total energy and its conversion into fracture energy with high fidelity. In contrast, the Hybrid DD model exhibits noticeable deviations, particularly near crack initiation, and shows discrepancies in both the timing and magnitude of energy dissipation. These differences indicate a loss of physical consistency when the model is applied to geometries not included in the training data, suggesting that the DD model may produce unreliable predictions outside the training range.

The phase-field evolution in Fig.~\ref{fig:val_geometry_field}(d) provides further insight into these differences. The Hybrid GNN--FEM (PI-DD) accurately captures the crack trajectory, including its deviation induced by the hole, and successfully predicts crack propagation not only from the initial sharp pre-crack but also from the circular hole. This behavior reflects the ability of the model to capture crack initiation and propagation driven by local stress concentrations. In contrast, the Hybrid DD model captures the overall crack direction and morphology reasonably well, but exhibits discrepancies in the progression of damage with respect to loading. In particular, the DD model tends to predict earlier crack evolution compared to both the Full FEM and PI-DD results.

The proposed framework retains predictive accuracy under geometric variations, including cases where crack initiation occurs at locations different from the initial pre-crack due to stress concentrations, demonstrating robust generalization beyond the training setting.

\FloatBarrier

\subsection{Generalization to unseen loading conditions}

\begin{figure}[hbt!]
\centering
\includegraphics[width=\linewidth]{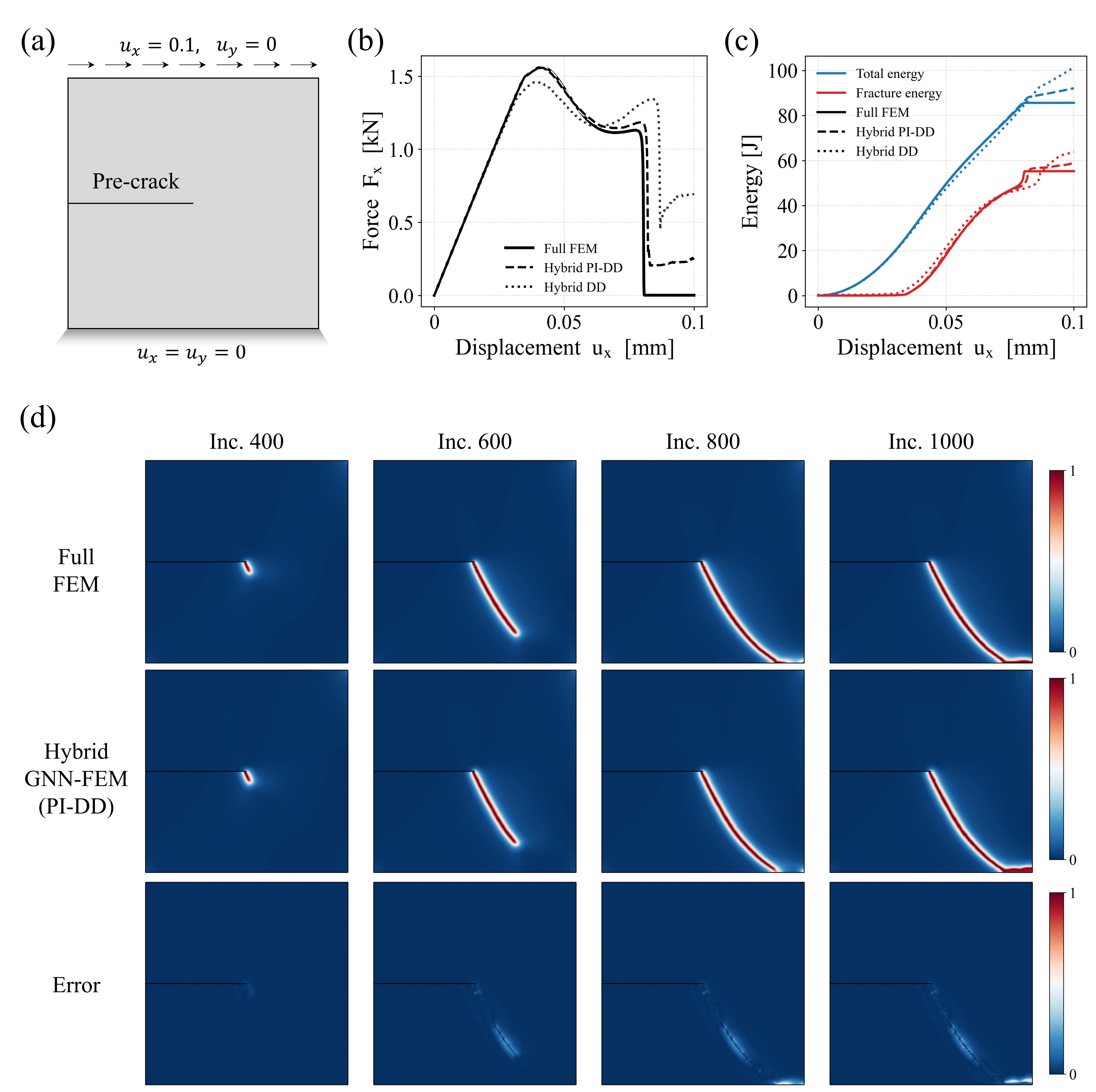}
\caption{Generalization under different loading conditions. 
(a) Problem setup with a modified loading condition. 
(b) Force--displacement responses comparing Full FEM, Hybrid GNN--FEM (PI-DD), and Hybrid DD. 
(c) Evolution of total and fracture energies. 
(d) Phase-field distributions at selected load increments for Full FEM, Hybrid GNN--FEM (PI-DD), and the corresponding error.}
\label{fig:val_loading_field}
\end{figure}

To examine the effect of loading variations, the trained GNN surrogate is applied 
to a problem with a modified loading direction and a different pre-crack location 
and length, as illustrated in Fig.~\ref{fig:val_loading_field}(a). Unlike the 
vertical loading condition used during training, shear loading is imposed in the 
horizontal direction.

Under this modified loading condition, the Hybrid GNN--FEM (PI-DD) model maintains 
close agreement with the Full FEM solution in both the force--displacement response 
and the overall energy evolution, as shown in Fig.~\ref{fig:val_loading_field}(b) 
and (c). The Hybrid GNN--FEM (DD) model, while capturing the general trend, shows 
noticeable deviations in the post-peak regime. Discrepancies between the total and 
fracture energies are observed in both the Full FEM and hybrid results, which 
originate from the hybrid phase-field formulation itself, as discussed in 
Section~\ref{subsec:pf_model}, and are not attributed to the surrogate model. 
Within this limitation, the PI-DD model follows the Full FEM trend more closely 
than the DD model, and the energy comparison is used here to assess whether the 
overall evolution trend is preserved.

Although the loading direction differs from that used during training, the hybrid 
framework remains applicable because the GNN surrogate operates on the history 
field $H$, which is computed by the FEM displacement solver at each load step. 
The effect of the changed loading condition is therefore reflected in the input 
to the surrogate through the updated history field, rather than being handled by 
the GNN directly. Consistent with the isotropic feature design of the GNN, which 
imposes no preferred crack orientation, Fig.~\ref{fig:val_loading_field}(d) shows 
that the model correctly captures the change in crack propagation direction induced 
by the shear loading, which deviates from the nearly straight crack path observed 
under the training configuration. A minor discrepancy is noted near the crack tip, 
where the phase-field does not fully reach $d = 1$, resulting in residual stiffness 
that introduces small deviations in the post-peak response.

\begin{figure}[hbt!]
\centering
\includegraphics[width=\linewidth]{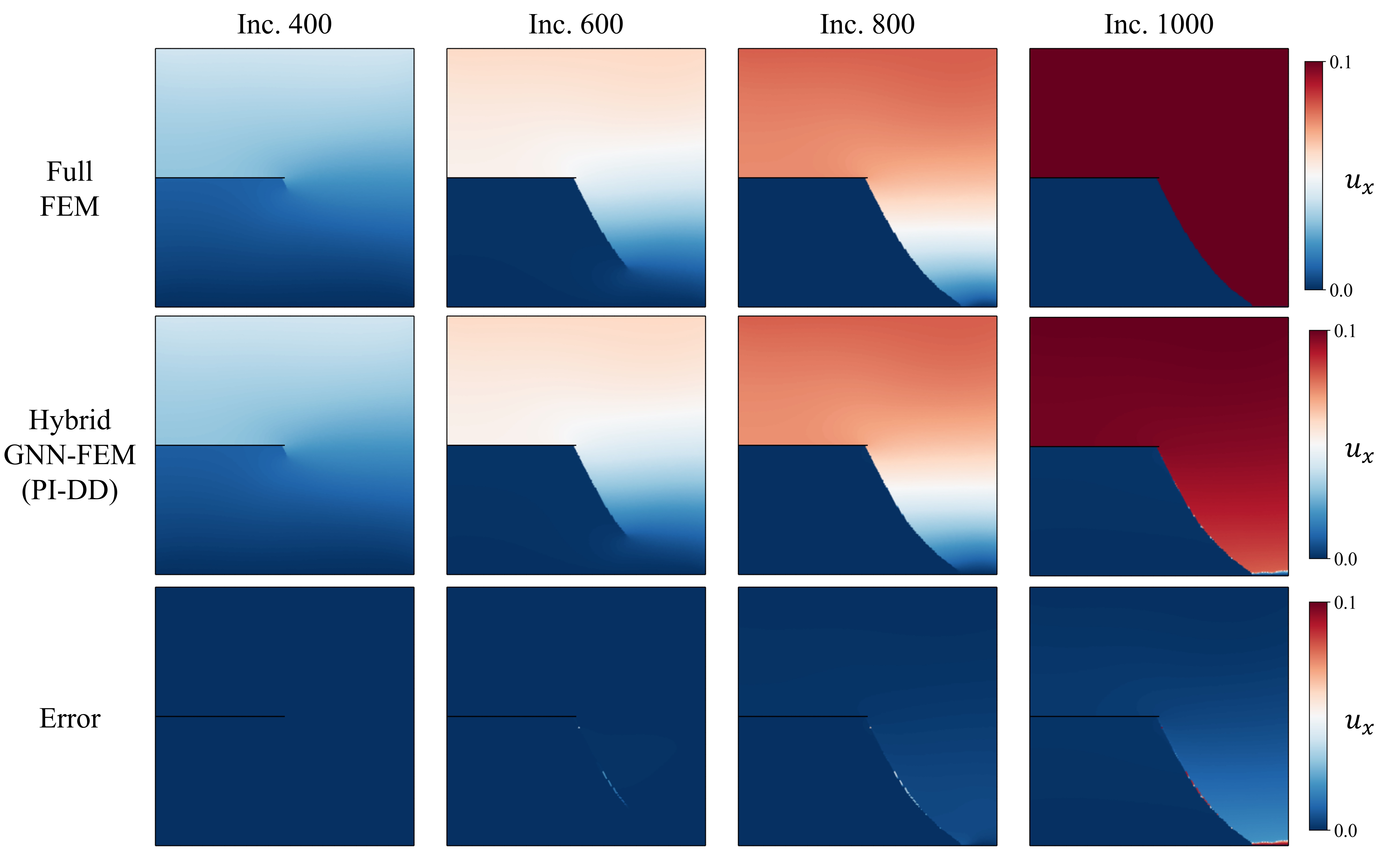}
\caption{Displacement field prediction under different loading conditions. 
Spatial distribution of the $u_x$ field at selected load increments (Inc. 400, 600, 800, and 1000) for Full FEM, Hybrid GNN--FEM (PI-DD), and the corresponding error.}
\label{fig:val_loading_ux}
\end{figure}

The displacement field comparison in Fig.~\ref{fig:val_loading_ux} further 
confirms that the predicted displacement field remains consistent with the Full 
FEM solution, as it is obtained directly from the finite element solve at each 
load step. The residual stiffness arising from the incomplete phase-field damage 
prevents the reaction force from fully reducing to zero, as reflected in the 
force--displacement curve in Fig.~\ref{fig:val_loading_field}(b). Nevertheless, 
this discrepancy remains localized, and the overall crack evolution is well 
preserved under the modified loading condition, demonstrating that the proposed 
framework generalizes to loading configurations outside the training setting.

\FloatBarrier

\FloatBarrier

\subsection{Generalization to unseen material properties}

\begin{figure}[hbt!]
\centering
\includegraphics[width=\linewidth]{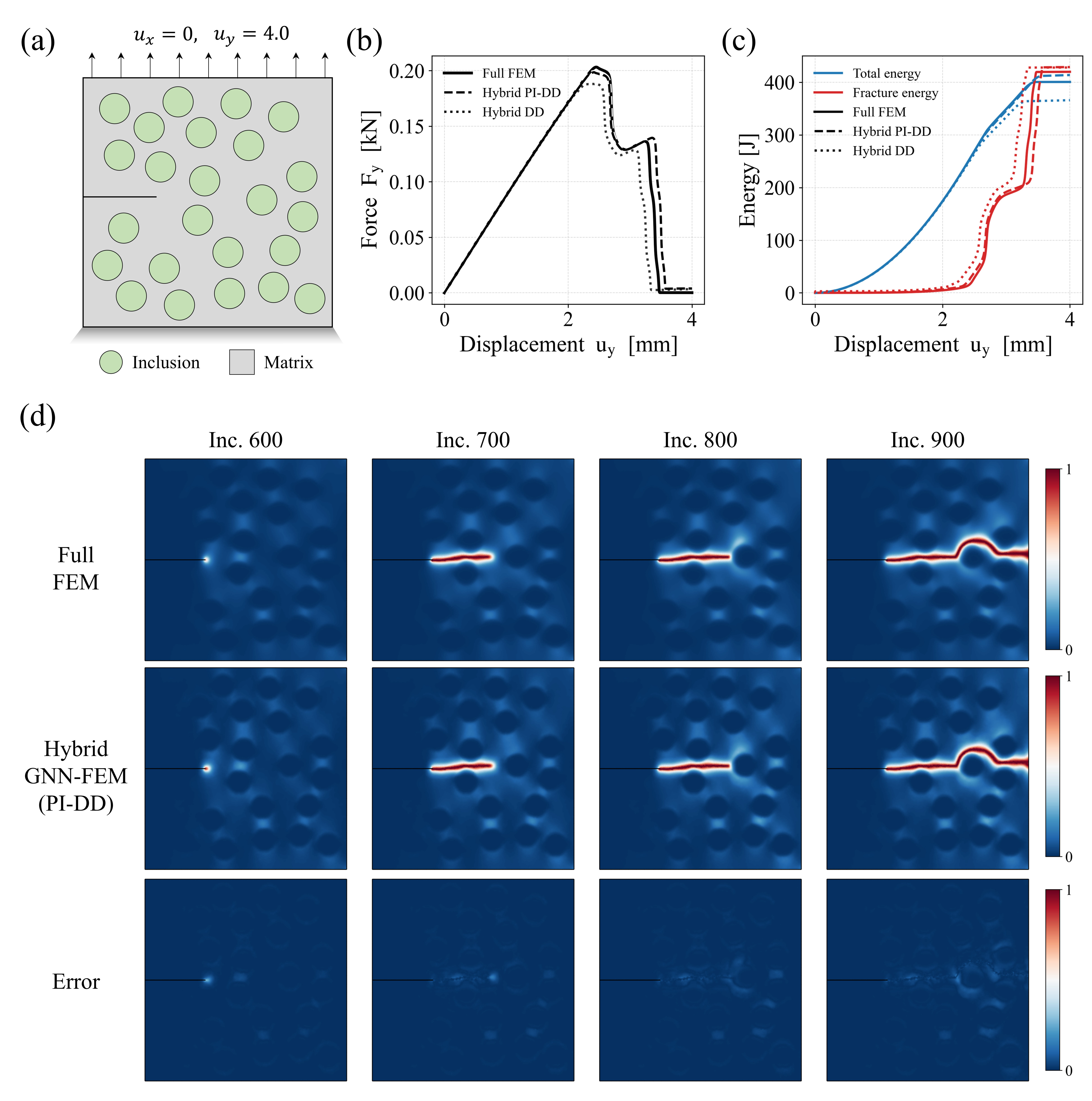}
\caption{Generalization to unseen material properties. 
(a) Problem setup with heterogeneous material properties, consisting of circular inclusions embedded in a matrix. 
(b) Force--displacement responses comparing Full FEM, Hybrid GNN--FEM (PI-DD), and Hybrid DD. 
(c) Evolution of total and fracture energies. 
(d) Phase-field distributions at selected load increments for Full FEM, Hybrid GNN--FEM (PI-DD), and the corresponding error.}
\label{fig:val_material_field}
\end{figure}

Finally, to assess generalization with respect to material properties, a heterogeneous geometry is considered in which circular inclusions are embedded within a softer matrix, as shown in Fig.~\ref{fig:val_material_field}(a). The material properties used in this case are summarized in Table~\ref{tab:material_properties}. Both the inclusion and matrix phases differ significantly from those used in training, particularly in terms of $E$ and $G_c$, with values differing by several orders. In addition, the material distribution is spatially heterogeneous, providing a challenging test for the model.

\setlength{\tabcolsep}{18pt}
\begin{table}[hbt!]
\centering
\caption{Material properties used for the heterogeneous validation case.}
\label{tab:material_properties}
\begin{tabular}{lccc}
\toprule
Material & $E$ [MPa] & $\nu$ & $G_c$ [N/mm] \\
\midrule
Inclusion & 2100 & 0.35 & 50.0 \\
Matrix    & 69   & 0.30 & 15.0 \\
\bottomrule
\end{tabular}
\end{table}

Despite these differences in material property scales, the Hybrid GNN--FEM (PI-DD) remains in close agreement with the Full FEM solution in both the force--displacement response and energy evolution, as shown in Fig.~\ref{fig:val_material_field}(b) and (c). The phase-field evolution in Fig.~\ref{fig:val_material_field}(d) shows that the PI-DD model captures crack propagation in the presence of material heterogeneity, including the interaction between the evolving crack and the inclusion phase. 
In particular, the predicted crack path propagates around the inclusions in close agreement with the Full FEM solution, capturing the influence of material heterogeneity on crack trajectory. The error remains localized, and the prediction retains high fidelity even under strong material contrast.

It is worth noting that this case involves simultaneous variations in multiple factors, including mesh resolution, load increments, pre-crack configuration, and material properties, with the latter differing in scale by several orders from those used during training. Despite these combined changes, the model produces consistent predictions without any additional training. This behavior is directly associated with the design choices adopted in the present work. In particular, the use of dimensionless features reduces sensitivity to material scales, while the physics-based residual terms and graph-based formulation allow the model to remain applicable under variations in discretization, geometry, and loading. As a result, the GNN model captures both spatial heterogeneity and variations in material properties, supporting the effectiveness of the proposed approach under complex and out-of-distribution scenarios.

\FloatBarrier

\section{Discussion}

The predictive behavior observed in the validation results can be interpreted in light of the challenges outlined in Section~\ref{subsec:challenges}. In phase-field fracture, the dominant difficulty stems from the combination of strong nonlinearity and history dependence, where the state at each load step is governed by the accumulated deformation history. Approximating such behavior through a direct mapping is inherently unstable, as small errors tend to propagate across increments, leading to deviations in the predicted crack path and, in some cases, qualitatively different fracture patterns. In the present framework, this issue is mitigated by retaining the incremental staggered solution procedure. The surrogate is restricted to the phase-field update, while the displacement and history fields continue to be resolved within the FEM loop, ensuring that the evolution remains tied to physically admissible states. The physics-informed loss further constrains the update through physics-based residual terms.

Additional challenges arise from variations in geometry, discretization, loading, and material properties, which alter the stress field and crack-driving behavior. Such variability makes generalization difficult for surrogate models that are tied to specific discretizations or problem configurations. The formulation adopted in this work simultaneously addresses these sources of variability. The graph-based representation allows the model to operate directly on unstructured meshes, accommodating changes in geometry and discretization without relying on a fixed mesh. At the same time, dimensionless features reduce sensitivity to absolute physical scales, which is essential when material properties or loading magnitudes vary. The physics-informed loss further stabilizes the response under conditions that differ from those used during training.

The framework is constructed by incorporating physical knowledge across feature design (dimensionless and physically meaningful inputs), loss formulation (physics-based residual terms), and model structure (graph-based representation and incremental coupling), guided by the challenges associated with nonlinear and history-dependent PDE systems. This design targets multiple sources of variability simultaneously, with the explicit objective of achieving generalization across varying problem settings. As a result, the formulation is not limited to reproducing solutions within the training range, but remains applicable under systematic variations in geometry, discretization, loading, and material properties.

From a numerical perspective, the framework can be interpreted as a targeted modification of the original solver. The displacement field, which enforces global equilibrium, is retained within FEM, whereas the phase-field equation—local in nature and computationally intensive—is replaced by the surrogate. This selective approximation reduces computational cost while preserving the structure of the solution procedure. The preservation of this staggered scheme plays a central role in maintaining stability, particularly in the presence of strong nonlinearity and path dependence.

Despite these observations, the proposed approach does not fully eliminate error accumulation inherent to incremental surrogate modeling. Although the incremental solution structure is preserved, the phase-field update is still approximated at each load step, and small discrepancies can propagate through the loading history, particularly in strongly nonlinear regimes where the phase-field evolution is highly sensitive to the local energy state. This behavior represents a limitation of approaches that replace part of a history-dependent evolution process with an approximate operator. The current formulation is also restricted to isotropic fracture; extension to anisotropic phase-field models would require inputs that encode material orientation, direction-dependent properties, and, depending on the formulation, additional damage or internal variables \cite{teichtmeister2017phase,bleyer2018phase}. Equivariant graph architectures may further improve generalization across rotated material orientations by enforcing coordinate-consistent representations \cite{satorras2021en}. Finally, because the physics-informed residual loss is imposed as a soft constraint, exact satisfaction of irreversibility, energy dissipation, or other thermodynamic admissibility conditions is not guaranteed. The computational acceleration is also bounded, since the surrogate is applied only to the phase-field evolution while the displacement field remains resolved through FEM.

In this sense, the current formulation reflects a balance between computational efficiency, stability, and generalizability. The surrogate is introduced in the phase-field update, which is more amenable to graph-based approximation, while the FEM displacement solve is retained to enforce global equilibrium and boundary conditions. Further acceleration could be pursued by extending surrogate modeling to the displacement field, for example using reduced-order models or neural operators such as DeepONet or Fourier neural operators. However, such approaches are inherently limited in generalization scope: rather than achieving full generalizability across arbitrary configurations, they are typically applicable to a restricted set of loading conditions, boundary conditions, or geometric configurations for which the surrogate is trained. Nevertheless, the present framework mitigates the impact of accumulated discrepancies by embedding the phase-field surrogate within a physics-based incremental solution procedure, thereby regulating error propagation in a physically consistent setting and providing a practical pathway toward improved generalization in nonlinear and history-dependent fracture problems.


\section{Conclusion}

In this work, a hybrid GNN--FEM framework has been developed for phase-field fracture, in which a GNN is integrated into a conventional staggered finite element procedure to approximate the phase-field evolution at each load increment while retaining the FEM-based displacement solver to enforce mechanical equilibrium. The formulation incorporates dimensionless feature design, a physics-based residual loss, and a graph-based architecture that operates directly on mesh-based domains, enabling the surrogate to function within the incremental solution structure without approximating the full solution trajectory. The proposed framework demonstrates accurate prediction of crack evolution and robust generalization across variations in geometry, mesh resolution, loading conditions, and material properties without requiring retraining, indicating that embedding data-driven components within the governing solution procedure provides a viable strategy for nonlinear and history-dependent fracture problems. More broadly, these results suggest that generalizable surrogate modeling for nonlinear PDE systems may depend less on brute-force data generation and more on identifying physically meaningful, scale-consistent, and incrementally structured learning targets. By combining this multi-level physics integration with selective data-driven approximation, the present framework provides a data-efficient pathway toward reliable scientific machine learning for phase-field fracture simulation.

Several directions remain for future work. The current formulation is limited to isotropic materials under quasi-static loading, and extending the framework to anisotropic material behavior will require corresponding developments in both feature design and surrogate architecture. Further gains in computational efficiency may be achieved by introducing surrogate models for the displacement field through reduced-order models or neural operators, although such extensions would involve a trade-off between generalizability and computational speed. Extension to three-dimensional fracture problems also represents a natural and practically important direction, which would require careful consideration of mesh complexity and training data generation.

\clearpage

\section*{Declaration of competing interest}
The authors declare that they have no known competing financial interests or personal relationships that could have appeared to influence the work reported in this paper

\section*{Data availability}
All data used in this study are available from the corresponding author upon reasonable request.

\section*{Acknowledgment}
This research was supported by the Ministry of Food and Drug Safety of Korea (Grant No. RS-2023-00215667) and by the InnoCORE Program of the Ministry of Science and ICT (MSIT), Republic of Korea (Grant No. N10260002).

\clearpage

\appendix

\renewcommand{\thesection}{Appendix \Alph{section}}

\renewcommand{\thefigure}{\Alph{section}\arabic{figure}}
\renewcommand{\thetable}{\Alph{section}\arabic{table}}
\renewcommand{\thealgorithm}{\Alph{section}\arabic{algorithm}}

\setcounter{figure}{0}
\setcounter{table}{0}
\setcounter{algorithm}{0}

\section{Generation of Spatially Correlated Random Field}

Spatially correlated random fields of $H/(G_c \ell_c)$ are generated to provide physically meaningful and spatially smooth inputs for the physics-informed training process. As shown in Algorithm~\ref{alg:random_field}, an initial Gaussian random field is first sampled, which serves as a stochastic basis without inherent spatial structure. To introduce spatial correlation, the sampled field is subsequently smoothed using a distance-based weighting scheme over local neighborhoods, effectively controlling the correlation length and suppressing high-frequency variations. A thresholding operation is then applied to distinguish regions of relatively high and low values, followed by a sigmoid transformation to ensure a smooth and continuous transition between these regions while preserving meaningful contrast in the field distribution. The resulting random field therefore exhibits both stochastic variability and controlled spatial coherence, making it suitable for representing heterogeneous distributions within the training framework. It should be noted that the specific procedure adopted here is not unique, and alternative random field generation approaches could also be employed within the same framework without affecting the overall methodology.

\begin{algorithm}[htb!]
\caption{Generation of spatially correlated random fields}
\label{alg:random_field}
\begin{algorithmic}[1]

\State \textbf{Input:} element coordinates $\mathbf{X}=\{(x_e,y_e)\}_{e=1}^{N}$, number of neighbors $k$, sigmoid sharpness $\kappa$, high-value fraction $\alpha$, scaling constant $C$
\State \textbf{Output:} random field $H/(G_c l_c)$
\State Sample Gaussian random field $z_e \sim \mathcal{N}(0,1)$ for $e=1,\dots,N$
\State Find $k$-nearest neighbors and distances $d_{e,j}$
\State $\ell \gets \mathrm{median}(d_{e,k})$
\State $w_{e,j} \gets \exp\!\left[-(d_{e,j}/\ell)^2\right]$
\State $\tilde{w}_{e,j} \gets \dfrac{w_{e,j}}{\sum_{j=1}^{k} w_{e,j}}$
\State $z_e^{(s)} \gets \sum_{j=1}^{k} \tilde{w}_{e,j} z_j$
\State Find $\tau$ such that 
\[
\frac{1}{N}\sum_{e=1}^{N}\mathbf{1}(z_e^{(s)}>\tau)=\alpha
\]
\State $p_e \gets \left(1+\exp[-\kappa(z_e^{(s)}-\tau)]\right)^{-1}$
\State $H/(G_c l_c) \gets {C\,p_e}$

\end{algorithmic}
\end{algorithm}

\FloatBarrier

\begin{table}[hbt!]
\centering
\caption{Parameters used for random field generation}
\label{tab:rf_param}
\begin{tabular}{p{2cm} p{2cm} p{9cm}}
\toprule
Parameter & Value & Description \\
\midrule
$k$ & 15 & Number of nearest neighbors for spatial smoothing \\
$\kappa$ & 50 & Sigmoid sharpness parameter \\
$\alpha$ & 0.01 & Fraction of high-value region \\
$C$ & 30 & Scaling constant  \\
\bottomrule
\end{tabular}
\end{table}
\FloatBarrier

\section{Geometry and boundary conditions of numerical validation cases}

\begin{figure}[htb!]
\centering
\includegraphics[width=\linewidth]{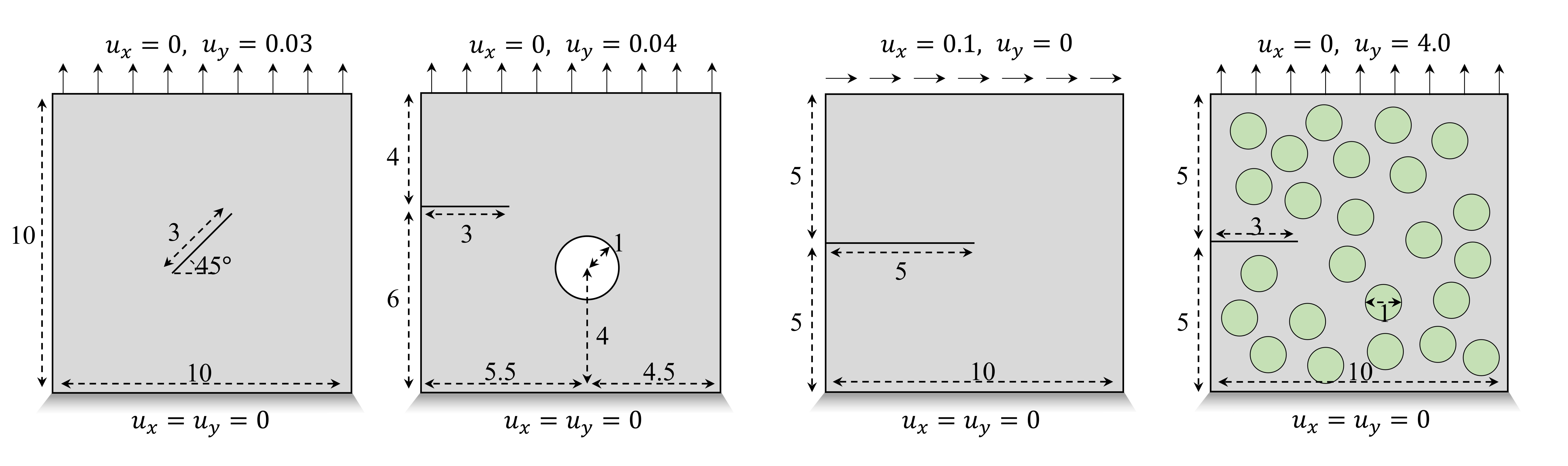}
\caption{Geometries and boundary conditions for the four numerical validation cases. 
From left to right, the configurations include a homogeneous square domain under vertical displacement loading, a domain containing a circular inclusion under non-uniform boundary conditions, a homogeneous domain subjected to horizontal displacement loading, and a heterogeneous domain with multiple circular inclusions. 
The bottom boundary is fixed, while prescribed displacements are applied at the top boundary.}
\label{fig:appendix_cases}
\end{figure}

\FloatBarrier

\clearpage

\begingroup
\setlength{\bibsep}{0pt}
\renewcommand{\baselinestretch}{0.9}\normalsize

\bibliography{references}

\endgroup

\end{document}